\documentclass[10pt,twocolumn,letterpaper]{article}

\usepackage[margin=0.75in]{geometry}
\usepackage{amsmath,amssymb,amsthm}
\usepackage{booktabs}
\usepackage{graphicx}
\usepackage{caption}
\usepackage{adjustbox}
\usepackage{multirow}
\usepackage{xcolor}
\usepackage{tikz}
\usetikzlibrary{arrows.meta,positioning,fit,calc,backgrounds,shapes.geometric}
\usepackage{pgfplots}
\usepackage{xspace}
\newcommand{\eg}{\emph{e.g.}\xspace}

\usepgfplotslibrary{groupplots}
\pgfplotsset{compat=1.16}

\usepackage[colorlinks=true,allcolors=blue]{hyperref}

\newtheorem{proposition}{Proposition}
\DeclareMathOperator*{\argmax}{arg\,max}
\newcommand{\norm}{\operatorname{norm}}
\newcommand{\Y}{\mathcal{Y}}
\newcommand{\C}{\mathcal{C}}

\definecolor{cBlue}{HTML}{0072B2}
\definecolor{cOrange}{HTML}{E69F00}
\definecolor{cGreen}{HTML}{009E73}
\definecolor{cVerm}{HTML}{D55E00}
\definecolor{cGrey}{HTML}{7F7F7F}
\definecolor{cSky}{HTML}{56B4E9}

\title{
Seeing Beyond Names: Image-Grounded Concept Linking\\
for Robust Vision-Language Recognition
}

\author{
Behraj Khan$^{1,3}$\thanks{Behraj Khan and Tahir Qasim Syed contributed equally and share first authorship.}
\quad
Tahir Qasim Syed$^{3}$\footnotemark[1]
\\
Shabir Ahmad$^{2}$
\quad
Syed Ahmad Chan Bukhari$^{2}$\thanks{Corresponding author: Behraj Khan.}
\\[5pt]
\small $^{1}$St. John's University, New York, USA
\\
\small $^{2}$Gachon University, Seongnam, Republic of Korea
\\
\small $^{3}$Institute of Business Administration Karachi, Pakistan
\\[5pt]
\small \texttt{behraj.khan@stjohns.edu}
}
\date{}

\begin{document}

\maketitle

\begin{abstract}
A zero-shot vision-language classifier is highly sensitive to the class names it receives. Even when images remain unchanged, replacing familiar labels with synonyms, abbreviations, or ontology terms can significantly alter predictions. We investigate this \emph{vocabulary shift} using a leakage-free evaluation protocol across Tiny ImageNet, ImageNet-R, ImageNet-A, and ImageNetV2. With CLIP ViT-B/16, alternative class names reduce top-1 accuracy by $14.8$--$17.6$ percentage points.
To address this issue, we propose \emph{image-grounded concept linking}, a training-free method that maps class names to visual concepts and uses a reliability gate to improve predictions. The method outperforms a template ensemble across all four datasets, achieving a pooled gain of $1.8$ points.
We further evaluate five vision-language backbones across four datasets. Our method improves performance in $16$ of $20$ backbone--dataset combinations, with no significantly negative results. It also reduces pooled negative log-likelihood across all five backbones. Stronger text linking and transductive refinement provide no consistent improvements, suggesting that accurate visual class-name linking remains the main challenge. A second CLIP ViT-B/16 seed confirms the improvement on ImageNetV2, but not on ImageNet-A.
\end{abstract}

\section{Introduction}
\begin{figure*}[t]
\centering
\begin{minipage}[b]{0.665\textwidth}
\centering
\adjustbox{max width=\linewidth}{\begin{tikzpicture}[font=\footnotesize\sffamily, >={Stealth[length=1.6mm]}, line join=round,
  lab/.style={font=\scriptsize\sffamily, align=center, text=black!80, anchor=north},
  sub/.style={font=\scriptsize\sffamily, text=black!60},
  arr/.style={->, draw=black!70, line width=0.6pt},
  grp/.style={draw=black!25, dashed, rounded corners=4pt, inner sep=1.6mm},
  snow/.pic={\foreach \a in {0,60,120}{\draw[cBlue, line width=0.45pt] (\a:0.1)--(\a+180:0.1);
             \foreach \s in {1,-1}{\draw[cBlue, line width=0.35pt] (\a:\s*0.06)--++(\a+\s*45:0.035); \draw[cBlue, line width=0.35pt] (\a:\s*0.06)--++(\a-\s*45:0.035);}}},
  enc/.pic={
     \draw[fill=cSky!22, draw=cBlue!80, line width=0.6pt, rounded corners=1pt] (-0.5,-0.5)--(-0.5,0.5)--(0.5,0.28)--(0.5,-0.28)--cycle;
     \foreach \x in {-0.3,-0.05,0.2}{\draw[cBlue!45, line width=0.4pt] (\x,{-0.43+0.22*(\x+0.5)})--(\x,{0.43-0.22*(\x+0.5)});}
     \path (0.5,0.5) pic {snow};}]
\def\yA{0} \def\yB{-2.25}
\coordinate (S)   at (0.75,\yA);  \coordinate (G)  at (2.85,\yA); \coordinate (L) at (5.05,\yA); \coordinate (T) at (7.55,\yA);
\coordinate (I)   at (0.75,\yB);  \coordinate (F)  at (2.85,\yB); \coordinate (V) at (7.55,\yB); \coordinate (Q) at (9.75,\yB); \coordinate (Y) at (11.45,\yB);
\begin{scope}[shift={(S)}]
  \draw[fill=cOrange!14, draw=cOrange!90!black, line width=0.6pt, rounded corners=1.5pt] (-0.72,-0.27)--(0.45,-0.27)--(0.72,0)--(0.45,0.27)--(-0.72,0.27)--cycle;
  \draw[fill=white, draw=cOrange!90!black, line width=0.5pt] (0.48,0) circle (0.055);
  \node[font=\scriptsize\itshape] at (-0.13,0) {bell toad};
  \node[lab] at (0,-0.42) {class string $x_y$};
\end{scope}
\path (G) pic {enc}; \node at (G) {$g$};
\node[lab] at ($(G)+(0,-0.62)$) {text encoder\\[-0.3ex]{\color{black!55}frozen}};
\begin{scope}[shift={(L)}]
  \foreach \y/\c in {-0.36/22,-0.12/16,0.12/10}{%
    \draw[fill=cOrange!\c, draw=cOrange!90!black, line width=0.55pt] (-0.42,\y+0.24)--(-0.42,\y) arc(180:360:0.42 and 0.11)--(0.42,\y+0.24);}
  \draw[fill=cOrange!6, draw=cOrange!90!black, line width=0.55pt] (0,0.36) ellipse (0.42 and 0.11);
  \foreach \x in {-0.2,0,0.2}{\fill[cOrange!90!black] (\x,0.36) circle (0.025);}
  \node[lab] at (0,-0.62) {lexicon memory $m_c$\\[-0.3ex]{\color{black!55}Eq.~\eqref{eq:memory}}};
\end{scope}
\begin{scope}[shift={(T)}]
  \draw[fill=white, draw=black!45, line width=0.5pt, rounded corners=2pt] (-0.82,-0.46) rectangle (0.82,0.46);
  \foreach \i/\len/\c/\n in {0/1.0/95/1,1/0.62/70/2,2/0.38/50/3,3/0.2/35/4}{%
    \node[font=\tiny, text=black!70, anchor=east, inner sep=0pt] at (-0.46,0.3-0.2*\i) {$c_{\n}$};
    \fill[cOrange!\c, rounded corners=0.6pt] (-0.4,0.25-0.2*\i) rectangle ++(\len*1.1,0.1);}
  \node[lab] at (0,-0.62) {top-$M$ concepts\\[-0.3ex]{\color{black!55}$P(c\mid x_y)$, Eq.~\eqref{eq:link}}};
\end{scope}
\begin{scope}[shift={(I)}]
  \draw[fill=cSky!12, draw=black!55, line width=0.6pt, rounded corners=1.5pt] (-0.6,-0.42) rectangle (0.6,0.42);
  \begin{scope}
    \clip[rounded corners=1.5pt] (-0.6,-0.42) rectangle (0.6,0.42);
    \fill[cOrange!80] (0.3,0.2) circle (0.1);
    \fill[cGreen!45] (-0.6,-0.42)--(-0.25,0.08)--(0.05,-0.18)--(0.32,0.02)--(0.6,-0.3)--(0.6,-0.42)--cycle;
    \fill[cGreen!80!black] (-0.6,-0.42)--(-0.35,-0.15)--(-0.05,-0.42)--cycle;
  \end{scope}
  \node[lab] at (0,-0.55) {test image $I$};
\end{scope}
\path (F) pic {enc}; \node at (F) {$\phi$};
\node[lab] at ($(F)+(0,-0.62)$) {image encoder\\[-0.3ex]{\color{black!55}frozen}};
\begin{scope}[shift={(V)}]
  \shade[inner color=white, outer color=cSky!60] (0,0) circle (0.46); \draw[cBlue!60, line width=0.5pt] (0,0) circle (0.46);
  \draw[black!35, line width=0.35pt] (-0.46,0) arc(180:360:0.46 and 0.13);
  \draw[black!25, dashed, line width=0.35pt] (0.46,0) arc(0:180:0.46 and 0.13);
  \foreach \p/\s/\o in {(-0.18,0.2)/1.0/0.75,(0.17,0.12)/0.75/0.55,(0.02,-0.22)/0.5/0.4}{%
    \fill[cOrange, opacity=\o] \p circle (0.11*\s);
    \fill[cOrange!90!black] \p circle (0.022);}
  \node[star, star points=5, star point ratio=2.2, fill=cGreen!80!black, inner sep=0pt, minimum size=5pt] at (0.12,-0.06) {};
  \node[lab] at (0,-0.55) {vMF mixture $\ell_y(I)$\\[-0.3ex]{\color{black!55}over $v_c$, Eq.~\eqref{eq:mix}}};
\end{scope}
\begin{scope}[shift={(Q)}]
  \draw[fill=black!4, draw=black!55, line width=0.55pt, rounded corners=2pt] (-0.5,-0.42) rectangle (0.5,0.42);
  \draw[black!35, line width=0.35pt] (-0.38,-0.28)--(0.38,-0.28);
  \draw[cVerm, line width=0.9pt, domain=-0.36:0.36, samples=30, smooth] plot (\x, {-0.26+0.5/(1+exp(-14*\x))});
  
  \node[lab] at (0,-0.55) {gate $\pi_y$\\[-0.3ex]{\color{black!55}Eq.~\eqref{eq:gate}}};
\end{scope}
\begin{scope}[shift={(Y)}]
  \draw[fill=cGreen!14, draw=cGreen!70!black, line width=0.6pt, rounded corners=2pt] (-0.36,-0.3) rectangle (0.36,0.3);
  \node at (0,0) {$\hat y$};
  \node[lab] at (0,-0.45) {prediction};
\end{scope}
\draw[arr] ($(S)+(0.76,0)$) -- ($(G)+(-0.55,0)$);
\draw[arr] ($(G)+(0.55,0)$) -- node[sub, above]{$h_y$} ($(L)+(-0.48,0)$);
\draw[arr] ($(L)+(0.48,0)$) -- ($(T)+(-0.86,0)$);
\draw[arr] ($(T)+(0,-1.2)$) -- node[sub, right, pos=0.28]{weights} ($(V)+(0,0.5)$);
\draw[arr] ($(I)+(0.64,0)$) -- ($(F)+(-0.55,0)$);
\draw[arr] ($(F)+(0.55,0)$) -- node[sub, above]{$\phi(I)$} ($(V)+(-0.5,0)$);
\draw[arr] ($(V)+(0.5,0)$) -- ($(Q)+(-0.54,0)$);
\draw[arr] ($(Q)+(0.54,0)$) -- ($(Y)+(-0.4,0)$);
\fill[black!70] ($(G)+(0.78,0)$) circle (0.03);
\draw[arr, dashed] ($(G)+(0.78,0)$) -- ($(G)+(0.78,0.98)$) -- node[sub, above, pos=0.5]{raw string $h_y$, kept as the gate's fallback component} (9.75,0.98) -- ($(Q)+(0,0.46)$);
\begin{scope}[on background layer]
  \node[grp, fit={($(S)+(-0.8,0.55)$) ($(T)+(1.0,-1.12)$)}] (G1) {};
  \node[grp, fit={($(I)+(-0.8,0.6)$) ($(Y)+(0.62,-1.12)$)}] (G2) {};
\end{scope}
\node[sub, anchor=south west] at (G1.north west) {once per class string, cached};
\node[sub, anchor=north west] at (G2.south west) {per test image: frozen encoders, no gradient steps};
\end{tikzpicture}
}\\[0.5mm]
{\small (a) Image-grounded concept linking}
\end{minipage}\hfill
\begin{minipage}[b]{0.315\textwidth}
\centering
\begin{tikzpicture}
\begin{axis}[font=\footnotesize\sffamily, width=0.95\linewidth, height=47mm, ybar, bar width=4.2pt,
  ymin=35, ymax=90, ytick={40,50,60,70,80,90}, ylabel={top-1 accuracy (\%)}, ylabel style={yshift=-1.5mm},
  symbolic x coords={Tiny-IN,IN-R,IN-A,IN-V2}, xtick=data, xticklabel style={font=\scriptsize\sffamily}, axis lines*=left, ymajorgrids, grid style={black!10},
  enlarge x limits=0.16, legend style={font=\scriptsize\sffamily, draw=none, fill=none, at={(0.5,1.02)}, anchor=south, legend columns=2, /tikz/every even column/.append style={column sep=4pt}},
  legend image code/.code={\draw[#1, draw=none] (0cm,-0.08cm) rectangle (0.22cm,0.12cm);}]
\addplot[fill=cGrey!55, draw=none] coordinates {(Tiny-IN,75.5) (IN-R,81.6) (IN-A,61.3) (IN-V2,70.8)}; \addlegendentry{dev.\ name}
\addplot[fill=cBlue!70, draw=none] coordinates {(Tiny-IN,60.8) (IN-R,65.1) (IN-A,43.8) (IN-V2,55.9)}; \addlegendentry{held-out, CLIP}
\addplot[fill=cOrange, draw=none] coordinates {(Tiny-IN,61.6) (IN-R,68.0) (IN-A,46.8) (IN-V2,58.9)}; \addlegendentry{held-out, Ours}
\addplot[fill=cGreen!70, draw=none] coordinates {(Tiny-IN,73.5) (IN-R,78.2) (IN-A,60.9) (IN-V2,70.7)}; \addlegendentry{oracle link}
\end{axis}
\end{tikzpicture}\\[0.5mm]
{\small (b) Accuracy before and after renaming}
\end{minipage}
\caption{\textbf{Vocabulary shift and image-grounded concept linking.} (a) A deployment string is encoded once and linked to a soft distribution over the top-$M$ concepts of a lexicon memory. For each test image, the class score is a von Mises--Fisher mixture over those candidates; a reliability gate adds the raw string as one more component when the link is weak, and the prediction is the highest gated score. Both encoders are frozen (snowflakes) and nothing is trained. The example string is illustrative. (b) CLIP ViT-B/16, seed 0, test classes of four datasets. Calling the classes by held-out WordNet lemmas instead of their development names costs 14.8--17.6 points, and a perfect linker would recover most of it. Ours closes part of the gap on every dataset (point estimates); Tab.~\ref{tab:diff} gives paired intervals against the template ensemble.}
\label{fig:teaser}
\end{figure*}
\label{sec:intro}

Contrastive vision-language models such as CLIP~\cite{radford2021clip} classify by matching. An image is compared with a text embedding of each class name, and the closest name wins. This is what makes them useful in open settings, and it is also a weak point, because the class name is an input and inputs vary. One hospital writes ``pleural effusion'', another ``PE'', and a third uses a term from its own ontology. A wildlife survey can use a common name or a Latin binomial. The images and labels are the same in every case; only the naming has changed.

There is evidence that this matters. CLIP's ImageNet-1k accuracy varies by about three points across three naming conventions for the same classes~\cite{thengane2022continual}. Many CLIP errors that are top-5 correct but top-1 wrong trace back to class names~\cite{ge2023improving}. Open-vocabulary detectors lose 8.8--9.5 AP$_{50}$ on novel classes when category names are replaced by synonyms~\cite{xie2026synclip}. Medical VLMs are sensitive to prompt wording~\cite{luo2026kepil,lai2024carzero}. Most fixes work on the text side, with better names~\cite{parisot2023name}, richer descriptions~\cite{menon2023dclip,pratt2023cupl}, hierarchies~\cite{novack2023chils,ge2023improving}, synonym spaces~\cite{yin2024s3}, or synonym-consistent pretraining~\cite{xie2026synclip}. They help when the developer controls the name. They do not answer a different question: \emph{at deployment, given a string the system has never seen, which concept does it mean?}

This is close to entity linking in NLP~\cite{liu2021sapbert} and to the ontology-agnostic tokenizers recently proposed for health records~\cite{ding2026mediator}. Vision has one advantage here: at test time we hold the image, and the image is evidence about which reading of a name is meant. Our method uses it (Fig.~\ref{fig:teaser}a). A lexicon proposes a short list of candidate concepts for each class string, and the image scores the class through a mixture over those candidates instead of one embedding fixed in advance. Candidate prototypes and image features are unit vectors, and CLIP's logit scale is a concentration, so the class score is exactly a von Mises--Fisher (vMF) mixture log-likelihood~\cite{banerjee2005vmf}. A gate adds the raw string as one more component when the link is weak.

We evaluate on four test sets and five frozen backbones with a protocol that holds the deployment names out of every component, and we report every comparison as a paired, class-level interval. The evidence supports five findings.
\begin{itemize}
\item \textbf{Vocabulary shift is large and consistent}: renaming costs 14.8--17.6 points, and perfect linking would recover 12.7--17.1 (Tab.~\ref{tab:main}).
\item \textbf{Our method helps, modestly}: with CLIP ViT-B/16 it beats a template ensemble on all four sets, by $+0.6$ to $+2.3$ points, and the pooled gain is $\mathbf{+1.8}$\,{\scriptsize$[+0.9,\,+2.7]$} (Tab.~\ref{tab:diff}, Fig.~\ref{fig:forest}). The reliability gate is the component that makes it work, on every backbone we tried.
\item \textbf{Better lexical linking does not transfer}: fusing CLIP with a sentence encoder raises top-1 linking of held-out lexicon strings from about 29\% to about 41\% on every set, yet it does not improve linking of class names, and it lowers accuracy when used for hard linking, on all 20 backbone--dataset pairs (Tab.~\ref{tab:link}, Fig.~\ref{fig:link}).
\item \textbf{Transductive link resolution is domain-dependent}: relative to an equally transductive CLIP baseline, it helps on Tiny ImageNet and ImageNetV2 and hurts on ImageNet-R and ImageNet-A (Tab.~\ref{tab:trans}). Across backbones its pooled effect is never significant.
\item \textbf{It carries over to other dual encoders, with smaller gains}: across five backbones and four sets, Ours $-$ CLIP ensemble is positive in 16 of 20 pairs and significantly negative in none. It is significant on IN-V2 for every backbone, and pooled over the sets it is positive for all five and significant for three (Tab.~\ref{tab:bb}, Fig.~\ref{fig:bb}).
\end{itemize}

\paragraph{Contributions.} (i) A definition of vocabulary shift, with metrics that separate accuracy, worst-vocabulary accuracy, all-vocabularies-correct rate, and consistency. (ii) A training-free, frozen-backbone classifier with a closed-form vMF-mixture score and a calibrated fallback. (iii) A leakage-checked protocol over four test sets and five frozen backbones, with fair inductive and transductive baselines. (iv) Two carefully controlled negative results that locate the bottleneck: the names of visual classes, not lexicon linking in general.

\section{Related Work} \label{sec:related}

\paragraph{Class names and prompts.} CLIP relies on carefully selected ImageNet class names~\cite{radford2021clip}. ~\cite{ge2023improving} show that abstract, overly specific, or inconsistent class names can cause errors and address them using WordNet hierarchies. Other approaches learn class-name embeddings from labelled images~\cite{parisot2023name}, expand synonyms~\cite{yin2024s3}, or study prompt and semantic variation~\cite{li2025mvp,yang2026certification}. SynCLIP~\cite{xie2026synclip} and prefix conditioning~\cite{saito2023prefix} modify pretraining. Unlike these methods, we keep the model frozen, use no labelled test-class images, and handle deployment names with unknown concept mappings.

\paragraph{Recognition without given names.} Some methods remove the need for predefined class names. CaSED~\cite{conti2023vic} retrieves candidate labels from image captions, while ~\cite{han2023whatsinaname} assign names to image clusters. Vocabulary-free few-shot learning~\cite{zanella2025vocabfree} uses generic prompts when target names are unavailable. In contrast, we assume a name is provided but its intended concept is unknown.

\paragraph{Richer class descriptions.} DCLIP~\cite{menon2023dclip} and CuPL~\cite{pratt2023cupl} use LLM-generated descriptions, while WaffleCLIP~\cite{roth2023waffle} explores random-word prompts. CHiLS~\cite{novack2023chils} expands classes into subclasses and aggregates their scores. Our approach is closest to CHiLS, but retrieves candidate concepts from a lexicon for an unfamiliar name and weights them by linking confidence.

\paragraph{Transduction for VLMs.} Transductive methods use unlabeled test images jointly. Martin~\cite{martin2024transductive} estimate class distributions using an EM-inspired objective, while TransCLIP~\cite{zanella2024transclip} regularizes text predictions with a KL penalty. Our transductive extension (Sec.~\ref{sec:ext}) instead re-estimates concept-linking weights and is compared against baselines using the same transductive setup.

\paragraph{Prompt learning, shift, and calibration.} CoOp~\cite{zhou2022coop}, CoCoOp~\cite{zhou2022cocoop}, MaPLe~\cite{khattak2023maple}, and TPT~\cite{shu2022tpt} adapt prompts to improve recognition. Other work studies distribution shifts~\cite{xiao2024anyshift,khan2025calshift} and prediction calibration~\cite{guo2017calibration,wang2024dac,wang2025dor}. Vocabulary shift differs because the image and label distributions remain unchanged; only the names used to represent classes change. Our reliability gate is related to calibration, as it reduces trust in lexicon-based predictions when a name poorly matches available concepts.

\paragraph{Dual-encoder backbones.} Beyond CLIP, related architectures include OpenCLIP~\cite{ilharco2021openclip}, trained on LAION~\cite{schuhmann2022laion,cherti2023scaling}, SigLIP~\cite{zhai2023siglip}, and EVA-CLIP~\cite{sun2023evaclip}. Since these models differ in training data and objectives, evaluating a method across multiple backbones is important. We therefore test our approach on five dual-encoder backbones (Sec.~\ref{sec:results}).

\paragraph{Concept normalization.} Concept normalization has been widely studied in biomedical applications. SapBERT~\cite{liu2021sapbert} links UMLS synonyms, while MEDIATOR~\cite{ding2026mediator} and MedTok~\cite{su2025medtok} represent clinical concepts using learned codes. KEPIL~\cite{luo2026kepil} and CARZero~\cite{lai2024carzero} improve medical VLM robustness through retraining. Sentence encoders~\cite{reimers2019sbert} can also improve word representations through contextual prompts, such as ``meaning: x''~\cite{ranjan2025oneword}. We adopt this strategy for our fused linker.

\section{Method}
\label{sec:method}

\paragraph{Vocabulary shift.}
Let $\mathcal D=\{(I_i,y_i)\}$ be labelled images with label set $\Y$. A \emph{naming function} $\nu:\Y\to\mathcal X$ gives each class a string. A frozen VLM with $\ell_2$-normalized encoders $\phi$ (image) and $g$ (text) classifies zero-shot by
\begin{equation}
p(y\mid I;\nu)\propto\exp\!\big(\langle\phi(I),\,g(\nu(y))\rangle/\tau\big),
\label{eq:zs}
\end{equation}
where $g(x)$ embeds ``a photo of a $\{x\}$.'' and $\tau=0.01$ is CLIP's temperature. Vocabulary shift is the case where $p(I,y)$ is fixed but the deployment naming $\nu_t$ differs from the development naming $\nu_s$. Given a family $\mathcal V$ of naming functions, we want high accuracy under every $\nu\in\mathcal V$, not just on average, and confidence that reflects how reliable the naming is.

\paragraph{Lexicon memory and soft linking.}
A lexicon is a set of concepts $\C$, each with surface strings (\emph{views}) $\mathcal X_c$; we use WordNet 3.0 noun synsets~\cite{miller1995wordnet}. Train views $\mathcal X^{\mathrm{tr}}_c$ may be used by every component, and held-out views only build test vocabularies. With templates $t=1,\dots,T$ and $h_{x,t}=g(\text{template}_t(x))$, each concept has a memory and a prototype,
\begin{equation}
\begin{gathered}
m_c=\norm\Big(\sum_{x\in\mathcal X^{\mathrm{tr}}_c}\tfrac1T\textstyle\sum_t h_{x,t}\Big),\\
v_c=\norm\big(g_c+\gamma\,\mu_c\big),\quad g_c=\norm\Big(\sum_{x\in\mathcal X^{\mathrm{tr}}_c}\bar h_x\Big),
\end{gathered}
\label{eq:memory}
\end{equation}
where $\bar h_x=\norm(\frac1T\sum_t h_{x,t})$, $\mu_c$ is the mean feature of a few training images of a seen class ($0$ otherwise), and $\gamma\ge0$. Nothing is trained. For a deployment string $x_y$ with query $h_y=g(x_y)$, let $\C_y$ be the $M$ concepts with the largest $\langle h_y,m_c\rangle$. The soft link is
\begin{equation}
P(c\mid x_y)=\frac{\exp(\langle h_y,m_c\rangle/\tau_\ell)}{\sum_{c'\in\C_y}\exp(\langle h_y,m_{c'}\rangle/\tau_\ell)},\quad c\in\C_y.
\label{eq:link}
\end{equation}

\paragraph{Image-grounded scoring and the gate.}
The linked score of class $y$ for image $I$ is
\begin{equation}
\ell_y(I)=\log\sum_{c\in\C_y}P(c\mid x_y)\,\exp\!\big(\langle\phi(I),v_c\rangle/\tau\big).
\label{eq:mix}
\end{equation}
Because $\|v_c\|=\|\phi(I)\|=1$, each term is proportional to a vMF density with mean $v_c$ and concentration $\kappa=1/\tau$. The normalizer depends only on $\kappa$ and the dimension, so Eq.~\eqref{eq:mix} is a mixture log-likelihood up to a class-independent constant, and a softmax over classes is the Bayes posterior under a uniform prior. The image picks the candidate it fits best, weighted by the text prior, instead of the classifier committing to the top-1 link. Eq.~\eqref{eq:zs} is the case where the only component is the string itself, and hard linking is $M=1$. With $\bar s_y=\max_c\langle h_y,m_c\rangle$ and $\pi_y=\sigma\big((\varepsilon-\bar s_y)/\rho\big)$, the gate adds the raw string as one more component:
\begin{equation}
s_y(I)=\log\!\Big[(1-\pi_y)\,e^{\ell_y(I)}+\pi_y\,e^{\langle\phi(I),h_y\rangle/\tau}\Big].
\label{eq:gate}
\end{equation}
Prediction is $\argmax_y s_y(I)$. Ties are broken by the identifier of the top-1 linked concept.

\begin{proposition}[what the classifier depends on]
\label{prop:inv}
For a fixed image, $s_y$ depends on $x_y$ only through $P(\cdot\mid x_y)$, $\pi_y$, and $h_y$; without the gate, only through $P(\cdot\mid x_y)$. Hence two naming functions that induce the same link distributions give identical ungated predictions and probabilities.
\end{proposition}
\noindent\emph{Proof.} Read off Eqs.~\eqref{eq:link}--\eqref{eq:gate}.\qed

\paragraph{Selection.} $(\gamma,M,\tau_\ell)$ are chosen by validation accuracy, and $(\varepsilon,\rho)$ by validation NLL, with $\varepsilon$ taken from the 5/10/20/30/50th percentiles of validation $\bar s_y$. Validation classes are disjoint from test classes. Class strings are linked once and cached. Per image, scoring costs one dot product with each of at most $|\Y|M$ prototypes.

\paragraph{Two extensions we test.}
\label{sec:ext}
\emph{(A) Fused text linking.} The pilot identified linking accuracy as the bottleneck, so we replace $\langle h_x,m_c\rangle$ by
\begin{equation}
\mathrm{sim}(x,c)=(1-\beta)\,\langle h_x,m_c\rangle+\beta\,\langle e_x,n_c\rangle,
\label{eq:fused}
\end{equation}
where $e_x$ is the all-mpnet-base-v2~\cite{reimers2019sbert} embedding of ``meaning: $x$''~\cite{ranjan2025oneword}. The concept memory $n_c=\norm(\mathrm{mean}_{x\in\mathcal X^{\mathrm{tr}}_c}e_x+\delta\,\hat e_c)$ optionally adds the embedding $\hat e_c$ of the WordNet gloss. Glosses are scrubbed of every held-out and out-of-lexicon string, and an assertion checks that none survives. To avoid choosing these settings on few classes:
\begin{itemize}
\item $(\beta,\delta)$ are chosen by top-1 link accuracy on 3000 held-out strings of \emph{non-class} physical-entity concepts.
\item The link temperature is chosen by NLL on the same strings.
\item The gate becomes a logistic model $P(\text{true concept}\in\text{top-}M\mid\bar s)$ fitted on those strings, with 10\% of them having their own concept masked to simulate out-of-lexicon names.
\end{itemize}
Only $(\gamma,M)$ are chosen on validation classes.

\emph{(B) Transductive link resolution.} When the unlabeled test pool $\{\phi_i\}_{i=1}^N$ is available, each class $y$ becomes a vMF mixture over slots $k$: its $M_t=\min(M,40)$ candidates and the gated raw string. The prior weights are $w^0_{yk}$, namely $(1-\pi_y)P(c\mid x_y)$ and $\pi_y$. EM re-estimates the weights by MAP with a Dirichlet prior,
\begin{equation}
w_{yk}\propto a\,w^0_{yk}+\sum_i r_{iy}\,\rho_{iyk},\qquad a=\alpha N/|\Y|,
\label{eq:em}
\end{equation}
where $r_{iy}$ are class posteriors and $\rho_{iyk}$ within-class slot posteriors. An optional term $\lambda\kappa\langle\phi_i,c_y\rangle$ adds the image centroid $c_y=\norm(\sum_i r_{iy}\phi_i)$; the class prior is fixed uniform, and EM runs for 10 iterations. For fairness, the transductive CLIP baselines run \emph{the same code} with one slot per class (the raw or template-mean embedding) and the same centroid term and grid. The difference between them and our variant therefore isolates the lexicon links.

\section{Experiments}
\label{sec:protocol}

\paragraph{Lexicon and held-out views.} We use all WordNet 3.0 noun synsets and drop lemmas with upper-case letters or digits (\eg Latin binomials). Class concepts own their strings, and each string belongs to one concept. For every concept with at least two views, $\min(2,\lfloor n_c/2\rfloor)$ views are held out at random, and the first lemma stays in training. Held-out views never enter the memories, the prototypes, the glosses, or any training; assertions check this.

\paragraph{Datasets.} We use Tiny ImageNet~\cite{le2015tiny}, ImageNet-R~\cite{hendrycks2021many}, ImageNet-A~\cite{hendrycks2021nae}, and ImageNetV2 (matched frequency)~\cite{recht2019imagenetv2}, all with WordNet class ids. Classes are split 30/70 into validation and test groups. Within each group, 10\% are \emph{out-of-lexicon} (OOL; all their views removed from the lexicon), and the rest are split into seen and unseen. A class is evaluated under shift if it has a held-out view or is OOL. This gives 65, 68, 66, and 294 test classes. Tiny ImageNet uses its training split for validation images and its official validation images for testing. The other three sets have no training split, so validation classes use their own images, with no class overlap with the test group; image prototypes are disabled there ($\gamma=0$, which was also selected on Tiny ImageNet).

\paragraph{Metrics and statistics.} Each test vocabulary draws one held-out view per class (OOL: any view), and we use three draws. We report:
\begin{itemize}
\item mean top-1 accuracy over draws and \emph{worst-vocabulary} accuracy;
\item the \emph{all-correct} rate, i.e. images correct under every draw;
\item \emph{consistency}, i.e. identical predictions across a pair of draws;
\item NLL and 15-bin ECE.
\end{itemize}
Differences are paired over classes, with 95\% bootstrap intervals (2000 resamples for accuracy, 1000 otherwise). For a cross-dataset summary we pool the four estimates by inverse variance with $\mathrm{SE}\approx(\mathrm{hi}-\mathrm{lo})/3.92$ and report $I^2$. The test sets share ImageNet classes, so this pooling ignores some dependence and its intervals are approximate.

\paragraph{Baselines.} Unless stated otherwise, all methods use frozen OpenAI CLIP ViT-B/16 (QuickGELU). The baselines are:
\begin{itemize}
\item CLIP with the raw deployment string, and with a three-template ensemble;
\item the concept of the nearest training string;
\item hard linking ($M=1$);
\item $k$-means codes ($|\C|/4$ text clusters, a same-encoder discrete control in the spirit of~\cite{ding2026mediator});
\item soft linking without the gate;
\item an oracle that links every string to its true concept;
\item in the transductive setting, CLIP raw and CLIP ensemble with the same EM refinement.
\end{itemize}

\paragraph{Backbones.} For the backbone study we load five models through OpenCLIP~\cite{ilharco2021openclip}: OpenAI CLIP ViT-B/16 and ViT-L/14~\cite{radford2021clip}, OpenCLIP ViT-B/16 trained on LAION-2B (\texttt{laion2b\_s34b\_b88k})~\cite{schuhmann2022laion,cherti2023scaling}, SigLIP ViT-B/16 (\texttt{webli})~\cite{zhai2023siglip}, and EVA02-CLIP B/16 (\texttt{merged2b\_s8b\_b131k}, LAION-2B plus COYO-700M)~\cite{sun2023evaclip}. The method has no backbone-specific part except the temperature, which we read from each model as $\tau=1/\exp(\text{logit scale})$: 0.0085 for SigLIP and 0.01 for the others. SigLIP also learns an additive logit bias. It is the same for every class, so it cancels in the softmax over classes, in Eq.~\eqref{eq:gate}, and in the EM responsibilities, and we drop it. SigLIP is trained with a sigmoid rather than a softmax loss, so its NLL and ECE under a class softmax should be read with that in mind; accuracy is unaffected. Per backbone we recompute all text and image features, the concept memories, the link calibration of Sec.~\ref{sec:ext}, and every selected hyper-parameter, using the same rules. The lexicon, the splits, and the vocabulary draws depend only on the seed, so they are identical across backbones; the sentence encoder of the fused linker is shared. The study covers all four test sets with seed 0 and the training-free arms. For CLIP B/16 we also ran seed 1, which draws a new class split, new held-out names and new vocabulary draws.

\section{Results}
\label{sec:results}

\begin{table}[!ht]
\centering\small
\setlength{\tabcolsep}{3.2pt}
\begin{adjustbox}{max width=\linewidth}\begin{tabular}{@{}lccccc@{}}
\toprule
Method & Tiny-IN & IN-R & IN-A & IN-V2 & Mean\\
\#test classes & 65 & 68 & 66 & 294 & \\
\midrule
\textcolor{cGrey}{CLIP, development name} & \textcolor{cGrey}{75.5} & \textcolor{cGrey}{81.6} & \textcolor{cGrey}{61.3} & \textcolor{cGrey}{70.8} & \textcolor{cGrey}{72.3}\\
\midrule
CLIP, raw name & 60.8 & 65.1 & 43.8 & 55.9 & 56.4\\
CLIP, template ensemble & 61.1 & 66.1 & 44.4 & 56.7 & 57.1\\
Nearest training string & 55.7 & 63.5 & 39.1 & 49.8 & 52.0\\
Hard link (memory) & 56.5 & 64.5 & 42.1 & 50.2 & 53.3\\
$k$-means codes & 48.5 & 56.5 & 40.0 & 41.6 & 46.7\\
Soft link, no gate & 58.9 & 65.6 & 45.0 & 53.6 & 55.8\\
\textbf{Ours} (soft link + gate) & \textbf{61.6} & \textbf{68.0} & \textbf{46.8} & \textbf{58.9} & \textbf{58.8}\\
\midrule
Fused hard link & 48.3 & 54.2 & 37.6 & 44.2 & 46.1\\
Ours + fused text linking & 61.1 & 65.8 & 46.7 & 58.0 & 57.9\\
\midrule
\textcolor{cGrey}{Oracle link (ceiling)} & \textcolor{cGrey}{73.5} & \textcolor{cGrey}{78.2} & \textcolor{cGrey}{60.9} & \textcolor{cGrey}{70.7} & \textcolor{cGrey}{70.8}\\
\bottomrule
\end{tabular}\end{adjustbox}
\caption{\textbf{Top-1 accuracy (\%) under held-out names}, inductive setting (one test image at a time). All methods use frozen CLIP ViT-B/16 and one seed. The first row (grey) uses each class's development name and is shown only as the unshifted reference. Bold: best non-oracle method per column. ``Fused'' rows are the text-linking extension of Sec.~\ref{sec:ext}.}
\label{tab:main}
\end{table}

\begin{table*}[!ht]
\centering\small
\setlength{\tabcolsep}{5pt}
\begin{adjustbox}{max width=\textwidth}\begin{tabular}{@{}lccccc@{}}
\toprule
Paired difference (points) & Tiny-IN ($n{=}65$) & IN-R ($n{=}68$) & IN-A ($n{=}66$) & IN-V2 ($n{=}294$) & Pooled$^\ddagger$\\
\midrule
Ours $-$ CLIP ens. & $+0.6$\,{\scriptsize$[-1.3,\,+2.8]$} & $+1.9$\,{\scriptsize$[-0.1,\,+4.0]$} & $\mathbf{+2.3}$\,{\scriptsize$[+0.2,\,+4.6]$} & $\mathbf{+2.2}$\,{\scriptsize$[+0.7,\,+3.6]$} & $\mathbf{+1.8}$\,{\scriptsize$[+0.9,\,+2.7]$}\\
Gate effect (Ours $-$ soft link) & $\mathbf{+2.8}$\,{\scriptsize$[+0.2,\,+5.6]$} & $\mathbf{+2.4}$\,{\scriptsize$[+0.8,\,+4.3]$} & $+1.8$\,{\scriptsize$[-1.1,\,+4.7]$} & $\mathbf{+5.2}$\,{\scriptsize$[+3.9,\,+6.6]$} & $\mathbf{+3.7}$\,{\scriptsize$[+2.8,\,+4.7]$}\\
Fused linking $-$ Ours & $-0.6$\,{\scriptsize$[-3.1,\,+1.8]$} & $-2.2$\,{\scriptsize$[-5.7,\,+0.7]$} & $-0.1$\,{\scriptsize$[-1.6,\,+1.4]$} & $-0.9$\,{\scriptsize$[-2.5,\,+0.9]$} & $-0.6$\,{\scriptsize$[-1.6,\,+0.4]$}\\
Fused hard $-$ hard link & $\mathbf{-8.1}$\,{\scriptsize$[-14.7,\,-1.2]$} & $\mathbf{-10.2}$\,{\scriptsize$[-16.7,\,-4.5]$} & $-4.5$\,{\scriptsize$[-9.7,\,+0.2]$} & $\mathbf{-6.0}$\,{\scriptsize$[-9.2,\,-3.1]$} & $\mathbf{-6.5}$\,{\scriptsize$[-8.7,\,-4.2]$}\\
Oracle $-$ CLIP raw & $\mathbf{+12.7}$\,{\scriptsize$[+6.1,\,+20.4]$} & $\mathbf{+13.1}$\,{\scriptsize$[+7.5,\,+19.1]$} & $\mathbf{+17.1}$\,{\scriptsize$[+11.6,\,+22.8]$} & $\mathbf{+14.9}$\,{\scriptsize$[+10.9,\,+19.0]$} & $\mathbf{+14.7}$\,{\scriptsize$[+12.1,\,+17.3]$}\\
\bottomrule
\end{tabular}\end{adjustbox}
\caption{\textbf{Paired accuracy differences} with 95\% class-level bootstrap intervals (2000 resamples). Bold: interval excludes zero. $^\ddagger$Inverse-variance pooling over the four test sets with $\mathrm{SE}\approx(\mathrm{hi}-\mathrm{lo})/3.92$. The test sets share ImageNet classes, so they are not independent, and the pooled intervals are approximate and likely too narrow.}
\label{tab:diff}
\end{table*}

\begin{figure*}[!ht]
\centering
\begin{tikzpicture}[font=\scriptsize\sffamily]
\begin{groupplot}[group style={group size=4 by 1, horizontal sep=4mm, yticklabels at=edge left},
  width=0.27\textwidth, height=34mm, ymin=0.4, ymax=5.6, ytick={1,2,3,4,5},
  yticklabels={pooled,IN-V2,IN-A,IN-R,Tiny-IN}, xmajorgrids, grid style={black!10}, axis lines*=left, xlabel={difference (points)}, clip=false]
\nextgroupplot[title={Ours $-$ CLIP ens.}, title style={font=\scriptsize\sffamily\bfseries, yshift=-1.5mm}, xmin=-1.75, xmax=5.07]
\draw[black!45, densely dashed] (axis cs:0,0.4) -- (axis cs:0,5.6);
\addplot[cBlue, mark=*, mark size=1.6pt, only marks, error bars/.cd, x dir=both, x explicit, error bar style={line width=0.7pt}] coordinates {(0.57,5) += (2.24,0) -= (1.85,0)};
\addplot[cOrange, mark=*, mark size=1.6pt, only marks, error bars/.cd, x dir=both, x explicit, error bar style={line width=0.7pt}] coordinates {(1.90,4) += (2.14,0) -= (2.00,0)};
\addplot[cGreen, mark=*, mark size=1.6pt, only marks, error bars/.cd, x dir=both, x explicit, error bar style={line width=0.7pt}] coordinates {(2.31,3) += (2.29,0) -= (2.09,0)};
\addplot[cVerm, mark=*, mark size=1.6pt, only marks, error bars/.cd, x dir=both, x explicit, error bar style={line width=0.7pt}] coordinates {(2.17,2) += (1.39,0) -= (1.51,0)};
\addplot[black, mark=diamond*, mark size=2.2pt, only marks, error bars/.cd, x dir=both, x explicit, error bar style={line width=0.7pt}] coordinates {(1.81,1) += (0.93,0) -= (0.93,0)};
\nextgroupplot[title={Gate effect}, title style={font=\scriptsize\sffamily\bfseries, yshift=-1.5mm}, xmin=-1.66, xmax=7.24]
\draw[black!45, densely dashed] (axis cs:0,0.4) -- (axis cs:0,5.6);
\addplot[cBlue, mark=*, mark size=1.6pt, only marks, error bars/.cd, x dir=both, x explicit, error bar style={line width=0.7pt}] coordinates {(2.77,5) += (2.84,0) -= (2.56,0)};
\addplot[cOrange, mark=*, mark size=1.6pt, only marks, error bars/.cd, x dir=both, x explicit, error bar style={line width=0.7pt}] coordinates {(2.37,4) += (1.89,0) -= (1.58,0)};
\addplot[cGreen, mark=*, mark size=1.6pt, only marks, error bars/.cd, x dir=both, x explicit, error bar style={line width=0.7pt}] coordinates {(1.78,3) += (2.89,0) -= (2.83,0)};
\addplot[cVerm, mark=*, mark size=1.6pt, only marks, error bars/.cd, x dir=both, x explicit, error bar style={line width=0.7pt}] coordinates {(5.25,2) += (1.38,0) -= (1.33,0)};
\addplot[black, mark=diamond*, mark size=2.2pt, only marks, error bars/.cd, x dir=both, x explicit, error bar style={line width=0.7pt}] coordinates {(3.74,1) += (0.94,0) -= (0.94,0)};
\nextgroupplot[title={Fused linking $-$ Ours}, title style={font=\scriptsize\sffamily\bfseries, yshift=-1.5mm}, xmin=-6.27, xmax=2.42]
\draw[black!45, densely dashed] (axis cs:0,0.4) -- (axis cs:0,5.6);
\addplot[cBlue, mark=*, mark size=1.6pt, only marks, error bars/.cd, x dir=both, x explicit, error bar style={line width=0.7pt}] coordinates {(-0.58,5) += (2.40,0) -= (2.52,0)};
\addplot[cOrange, mark=*, mark size=1.6pt, only marks, error bars/.cd, x dir=both, x explicit, error bar style={line width=0.7pt}] coordinates {(-2.21,4) += (2.91,0) -= (3.46,0)};
\addplot[cGreen, mark=*, mark size=1.6pt, only marks, error bars/.cd, x dir=both, x explicit, error bar style={line width=0.7pt}] coordinates {(-0.06,3) += (1.43,0) -= (1.52,0)};
\addplot[cVerm, mark=*, mark size=1.6pt, only marks, error bars/.cd, x dir=both, x explicit, error bar style={line width=0.7pt}] coordinates {(-0.86,2) += (1.79,0) -= (1.68,0)};
\addplot[black, mark=diamond*, mark size=2.2pt, only marks, error bars/.cd, x dir=both, x explicit, error bar style={line width=0.7pt}] coordinates {(-0.59,1) += (0.97,0) -= (0.97,0)};
\nextgroupplot[title={Link EM $-$ transd.\ CLIP ens.}, title style={font=\scriptsize\sffamily\bfseries, yshift=-1.5mm}, xmin=-9.73, xmax=9.00]
\draw[black!45, densely dashed] (axis cs:0,0.4) -- (axis cs:0,5.6);
\addplot[cBlue, mark=*, mark size=1.6pt, only marks, error bars/.cd, x dir=both, x explicit, error bar style={line width=0.7pt}] coordinates {(3.31,5) += (4.40,0) -= (3.58,0)};
\addplot[cOrange, mark=*, mark size=1.6pt, only marks, error bars/.cd, x dir=both, x explicit, error bar style={line width=0.7pt}] coordinates {(-4.22,4) += (3.35,0) -= (4.22,0)};
\addplot[cGreen, mark=*, mark size=1.6pt, only marks, error bars/.cd, x dir=both, x explicit, error bar style={line width=0.7pt}] coordinates {(-3.73,3) += (3.74,0) -= (3.48,0)};
\addplot[cVerm, mark=*, mark size=1.6pt, only marks, error bars/.cd, x dir=both, x explicit, error bar style={line width=0.7pt}] coordinates {(2.22,2) += (2.41,0) -= (2.37,0)};
\addplot[black, mark=diamond*, mark size=2.2pt, only marks, error bars/.cd, x dir=both, x explicit, error bar style={line width=0.7pt}] coordinates {(0.04,1) += (1.61,0) -= (1.61,0)};
\end{groupplot}
\end{tikzpicture}
\caption{\textbf{Paired differences per test set} (coloured; 95\% class-level bootstrap intervals) and pooled (black diamond; approximate). Dashed line: no difference. Ours beats the CLIP ensemble on every set, and the gate is positive on every set. Fused linking never helps, and transductive link resolution changes sign across domains.}
\label{fig:forest}
\end{figure*}
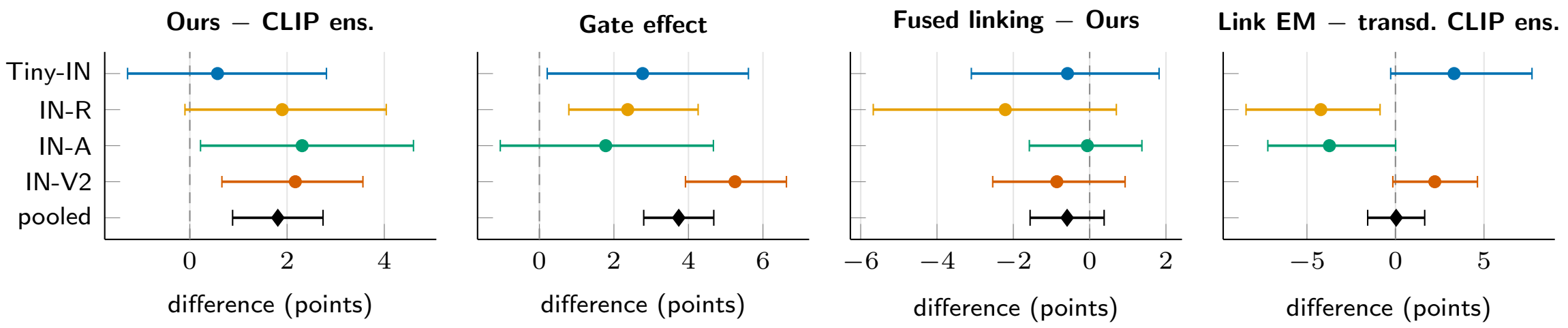

\paragraph{Vocabulary shift is large and consistent.} With development names, raw CLIP reaches 75.5, 81.6, 61.3, and 70.8\% on the four test sets. With held-out names it reaches 60.8, 65.1, 43.8, and 55.9\%, a loss of 14.8--17.6 points (Tab.~\ref{tab:main}, Fig.~\ref{fig:teaser}b). The oracle link recovers $\mathbf{+14.7}$\,{\scriptsize$[+12.1,\,+17.3]$} points when pooled ($I^2=0.00$) and stays $\mathbf{+12.9}$\,{\scriptsize$[+10.3,\,+15.6]$} points above the fused variant, so linking has large headroom.

\paragraph{Our method helps, and the gate is why.} Ours is the best non-oracle inductive method on all four sets (Tab.~\ref{tab:main}). Against the template ensemble the gain is positive on all four sets and significant on IN-A ($\mathbf{+2.3}$\,{\scriptsize$[+0.2,\,+4.6]$}) and IN-V2 ($\mathbf{+2.2}$\,{\scriptsize$[+0.7,\,+3.6]$}). The pooled estimate is $\mathbf{+1.8}$\,{\scriptsize$[+0.9,\,+2.7]$} with no measurable heterogeneity ($I^2=0.00$; Tab.~\ref{tab:diff}, Fig.~\ref{fig:forest}). Soft linking alone is below the ensemble on three of four sets (58.9 vs.\ 61.1, 65.6 vs.\ 66.1, 53.6 vs.\ 56.7). Adding the gate raises it by $\mathbf{+3.7}$\,{\scriptsize$[+2.8,\,+4.7]$} points pooled, with the per-set effect positive everywhere and significant on three sets. Hard linking is worse than raw CLIP everywhere: committing to one concept before seeing the image is costly.

\paragraph{Robustness and calibration (supplementary Tabs.~\ref{tab:full}, \ref{tab:extra}).} Compared with the ensemble, Ours has a higher all-correct rate on every set, significantly on IN-A ($\mathbf{+2.5}$\,{\scriptsize$[+0.4,\,+4.8]$}) and IN-V2 ($\mathbf{+2.2}$\,{\scriptsize$[+0.6,\,+3.6]$}). Its consistency is higher on Tiny-IN ($\mathbf{+1.3}$\,{\scriptsize$[+0.4,\,+2.5]$}). Its NLL and ECE are lower on every set in point estimate, significantly on IN-A (NLL $\mathbf{-0.116}$\,{\scriptsize$[-0.208,\,-0.033]$} nats, ECE $\mathbf{-3.6}$\,{\scriptsize$[-5.0,\,-2.1]$} points).

\begin{table}[!ht]
\centering\small
\setlength{\tabcolsep}{2.6pt}
\begin{adjustbox}{max width=\linewidth}\begin{tabular}{@{}llccc@{}}
\toprule
& Held-out strings & NN string & Memory & Fused\\
\midrule
Tiny-IN & all lexicon ($n{=}16{,}449$) & 29.2 / 46.3 & 28.9 / 46.7 & \textbf{40.8} / 62.2\\
 & test-class ($n{=}64$) & \textbf{45.3} / 62.5 & 43.8 / 64.1 & 40.6 / 64.1\\
\addlinespace[2pt]
IN-R & all lexicon ($n{=}16{,}458$) & 29.2 / 46.5 & 28.9 / 46.5 & \textbf{40.7} / 62.0\\
 & test-class ($n{=}67$) & 58.2 / 70.2 & \textbf{59.7} / 67.2 & 47.8 / 61.2\\
\addlinespace[2pt]
IN-A & all lexicon ($n{=}16{,}459$) & 29.2 / 46.5 & 29.1 / 46.7 & \textbf{40.5} / 61.7\\
 & test-class ($n{=}57$) & 45.6 / 57.9 & \textbf{49.1} / 61.4 & 49.1 / 61.4\\
\addlinespace[2pt]
IN-V2 & all lexicon ($n{=}16{,}414$) & 29.7 / 46.6 & 29.4 / 46.9 & \textbf{41.2} / 62.1\\
 & test-class ($n{=}268$) & \textbf{48.5} / 62.7 & 46.6 / 61.9 & 45.5 / 64.9\\
\bottomrule
\end{tabular}\end{adjustbox}
\caption{\textbf{Text-only linking}: concept-level top-1 / top-5 accuracy (\%) of held-out strings. The 3000 calibration strings are excluded. Fusion helps on lexicon strings by 11--12 points at top-1, but not on the names of test classes. Bold: best top-1 per row.}
\label{tab:link}
\end{table}

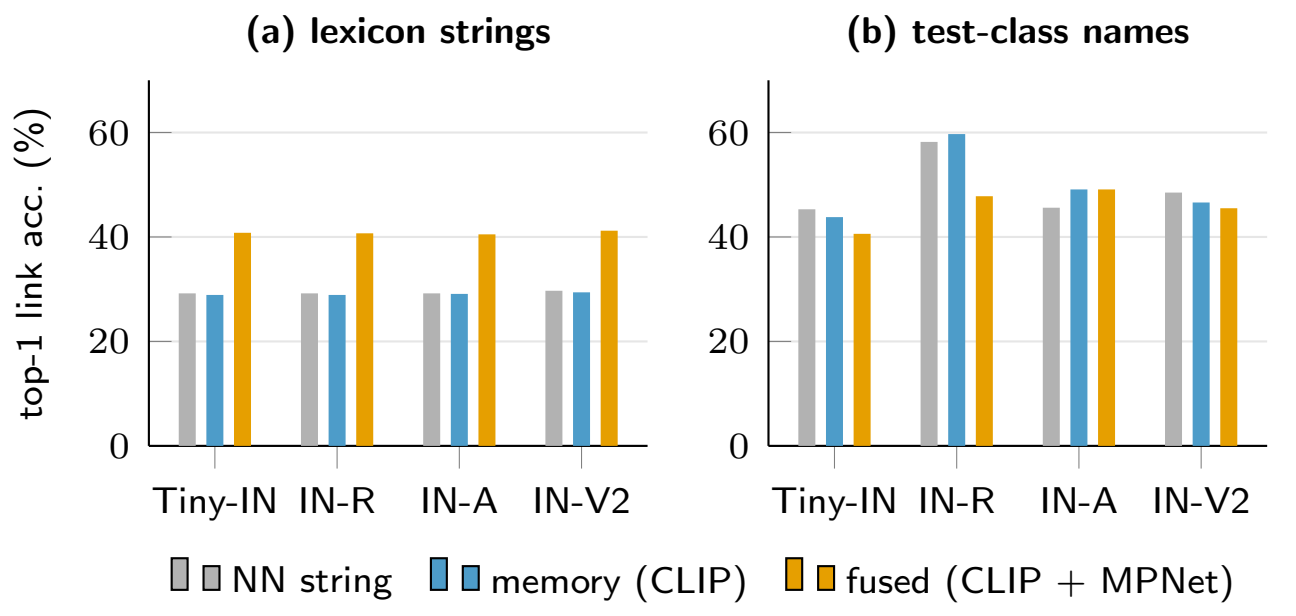
\begin{figure}[!ht]
\centering
\begin{tikzpicture}[font=\scriptsize\sffamily]
\begin{groupplot}[group style={group size=2 by 1, horizontal sep=8mm}, width=0.56\linewidth, height=40mm,
  axis lines*=left, ymajorgrids, grid style={black!10}, enlarge x limits=0.18, xticklabel style={font=\scriptsize\sffamily},
  legend image code/.code={\draw[#1, draw=none] (0cm,-0.08cm) rectangle (0.22cm,0.12cm);}]
\nextgroupplot[ylabel={top-1 link acc.\ (\%)}, title={(a) lexicon strings}, title style={font=\scriptsize\sffamily\bfseries, yshift=-1.5mm}, ybar, bar width=3.2pt, ymin=0, ymax=70, symbolic x coords={Tiny-IN,IN-R,IN-A,IN-V2}, xtick=data, legend to name=linkleg, legend columns=3, legend style={font=\scriptsize\sffamily, draw=none, /tikz/every even column/.append style={column sep=5pt}}]
\addplot[fill=cGrey!60, draw=none] coordinates {(Tiny-IN,29.2) (IN-R,29.2) (IN-A,29.2) (IN-V2,29.7)}; \addlegendentry{NN string}
\addplot[fill=cBlue!70, draw=none] coordinates {(Tiny-IN,28.9) (IN-R,28.9) (IN-A,29.1) (IN-V2,29.4)}; \addlegendentry{memory (CLIP)}
\addplot[fill=cOrange, draw=none] coordinates {(Tiny-IN,40.8) (IN-R,40.7) (IN-A,40.5) (IN-V2,41.2)}; \addlegendentry{fused (CLIP + MPNet)}
\nextgroupplot[title={(b) test-class names}, title style={font=\scriptsize\sffamily\bfseries, yshift=-1.5mm}, ybar, bar width=3.2pt, ymin=0, ymax=70, symbolic x coords={Tiny-IN,IN-R,IN-A,IN-V2}, xtick=data]
\addplot[fill=cGrey!60, draw=none] coordinates {(Tiny-IN,45.3) (IN-R,58.2) (IN-A,45.6) (IN-V2,48.5)};
\addplot[fill=cBlue!70, draw=none] coordinates {(Tiny-IN,43.8) (IN-R,59.7) (IN-A,49.1) (IN-V2,46.6)};
\addplot[fill=cOrange, draw=none] coordinates {(Tiny-IN,40.6) (IN-R,47.8) (IN-A,49.1) (IN-V2,45.5)};
\end{groupplot}
\node at ($(group c1r1.south)!0.5!(group c2r1.south)+(0,-9mm)$) {\pgfplotslegendfromname{linkleg}};
\end{tikzpicture}
\caption{\textbf{Lexicon linking does not transfer to class names.} Fusing CLIP text with a sentence encoder (Eq.~\eqref{eq:fused}) raises top-1 linking on $\approx$16k held-out lexicon strings from $\approx$29\% to $\approx$41\% on every set (left), but not on the held-out names of test classes (right; $n=57$--$268$).}
\label{fig:link}
\end{figure}

\paragraph{Better lexical linking does not transfer.} The fused linker of Sec.~\ref{sec:ext}(A) selected $\beta=0.5$ on every set, with $\delta=0.5$ on three of them (Tab.~\ref{tab:sel}). It raises top-1 linking of held-out lexicon strings by 11--12 points and top-5 by about 15 points on all four sets (Tab.~\ref{tab:link}, Fig.~\ref{fig:link}). On the held-out names of the \emph{test classes}, however, it is not better than the CLIP memory, and on IN-R it is clearly worse (47.8 vs.\ 59.7\%). In classification, fused hard linking loses to CLIP-memory hard linking on every set ($\mathbf{-6.5}$\,{\scriptsize$[-8.7,\,-4.2]$} pooled), and the fused gated variant does not beat Ours ($-0.6$\,{\scriptsize$[-1.6,\,+0.4]$} pooled). The calibration strings are held-out names of physical objects that are not classes, and the gain on them does not carry over to the visually defined concepts that make up ImageNet classes. Optimizing linking on a generic lexicon is therefore not a reliable proxy for optimizing it on class names.

\begin{table*}[!ht]
\centering\small
\setlength{\tabcolsep}{4.5pt}
\begin{adjustbox}{max width=\textwidth}\begin{tabular}{@{}lcccccccc@{}}
\toprule
 & \multicolumn{2}{c}{Tiny-IN} & \multicolumn{2}{c}{IN-R} & \multicolumn{2}{c}{IN-A} & \multicolumn{2}{c}{IN-V2}\\
\cmidrule(lr){2-3}\cmidrule(lr){4-5}\cmidrule(lr){6-7}\cmidrule(lr){8-9}
Transductive method & Acc & ECE & Acc & ECE & Acc & ECE & Acc & ECE\\
\midrule
CLIP raw + transduction & 63.5 & 19.9 & 67.3 & 19.4 & 48.9 & 33.8 & 58.4 & 36.6\\
CLIP ens.\ + transduction & 63.8 & 19.7 & 67.8 & 19.0 & 49.1 & 25.2 & 58.7 & 36.6\\
Link EM only ($\lambda{=}0$) & 63.3 & 4.2 & 66.1 & 1.9 & 45.9 & 7.0 & 57.7 & 4.3\\
Link EM + centroids & 67.2 & 16.1 & 63.6 & 23.5 & 45.4 & 35.8 & 60.9 & 27.3\\
\midrule
\multicolumn{9}{@{}l}{\emph{Paired accuracy differences (points), 95\% CI}}\\
CLIP ens.\ + transd.\ $-$ CLIP ens. & \multicolumn{2}{c}{$\mathbf{+2.8}$\,{\scriptsize$[+1.5,\,+4.1]$}} & \multicolumn{2}{c}{$\mathbf{+1.8}$\,{\scriptsize$[+0.1,\,+3.5]$}} & \multicolumn{2}{c}{$\mathbf{+4.7}$\,{\scriptsize$[+2.3,\,+7.1]$}} & \multicolumn{2}{c}{$\mathbf{+1.9}$\,{\scriptsize$[+0.5,\,+3.4]$}}\\
Link EM + centroids $-$ CLIP ens.\ + transd. & \multicolumn{2}{c}{$+3.3$\,{\scriptsize$[-0.3,\,+7.7]$}} & \multicolumn{2}{c}{$\mathbf{-4.2}$\,{\scriptsize$[-8.4,\,-0.9]$}} & \multicolumn{2}{c}{$-3.7$\,{\scriptsize$[-7.2,\,0.0]$}} & \multicolumn{2}{c}{$+2.2$\,{\scriptsize$[-0.1,\,+4.6]$}}\\
Link EM only $-$ fused inductive & \multicolumn{2}{c}{$\mathbf{+2.2}$\,{\scriptsize$[+0.1,\,+4.8]$}} & \multicolumn{2}{c}{$+0.3$\,{\scriptsize$[-0.8,\,+1.5]$}} & \multicolumn{2}{c}{$-0.8$\,{\scriptsize$[-2.7,\,+1.0]$}} & \multicolumn{2}{c}{$-0.3$\,{\scriptsize$[-1.5,\,+0.9]$}}\\
\bottomrule
\end{tabular}\end{adjustbox}
\caption{\textbf{Transductive setting} (the whole unlabeled test pool is available). Accuracy and ECE in \%. Both CLIP baselines use the same EM code and centroid refinement as our variant, with one slot per class, so the second difference row isolates the effect of the lexicon links. Pooled over the four sets: transduction alone $\approx$ $\mathbf{+2.5}$\,{\scriptsize$[+1.7,\,+3.3]$}; link EM + centroids $-$ CLIP ens.\ + transd.\ $\approx$ $0.0$\,{\scriptsize$[-1.6,\,+1.7]$} (heterogeneity $I^2=0.80$).}
\label{tab:trans}
\end{table*}

\paragraph{Transductive link resolution is domain-dependent.} Transduction alone helps CLIP on every set (CLIP ens.\ + transduction: $\mathbf{+2.5}$\,{\scriptsize$[+1.7,\,+3.3]$} pooled, Tab.~\ref{tab:trans}), in line with~\cite{zanella2024transclip,martin2024transductive}. Our transductive variant adds a further $+3.3$ and $+2.2$ points on Tiny-IN and IN-V2, where the intervals reach just below zero. It loses $-4.2$ and $-3.7$ points on IN-R and IN-A, and the IN-R interval excludes zero. The pooled effect is near zero with high heterogeneity ($I^2=0.80$). The link EM alone ($\lambda=0$) changes little, except on Tiny-IN ($\mathbf{+2.2}$\,{\scriptsize$[+0.1,\,+4.8]$}; pooled $+0.1$\,{\scriptsize$[-0.7,\,+0.8]$}). The sign changes come from combining it with the image centroids (link EM + centroids $-$ link EM: $+3.9$, $-2.5$, $-0.5$, $+3.2$ points). On renditions and adversarial images the centroid of a class's images is a poor anchor, and it can pull the weights toward the wrong candidate. Every transductive variant with a centroid term is poorly calibrated (ECE 16--37\%, baselines included), while link EM alone keeps ECE near the inductive level.

\paragraph{Selection and $M$.} In earlier runs, choosing $M$ for an uncalibrated soft link on 24 validation classes changed test accuracy by 2.8 points (60.8\% at $M=80$ vs.\ 58.0\% at $M=320$, same seed and splits). With the temperature and gate calibrated on 3000 strings, the fused variant is flat in $M$ on every set (supplementary Fig.~\ref{fig:msens}). For Ours, the validation-selected $M$ and $\tau_\ell$ still vary widely across sets (Tab.~\ref{tab:sel}). In most transductive runs the largest centroid weight ($\lambda=0.5$) was selected, for the baselines and for our variant, so the transductive comparison could change with wider grids.

\begin{table*}[!ht]
\centering\small
\setlength{\tabcolsep}{5pt}
\begin{adjustbox}{max width=\textwidth}\begin{tabular}{@{}llccccc@{}}
\toprule
Backbone & Pretraining data & Tiny-IN ($n{=}65$) & IN-R ($n{=}68$) & IN-A ($n{=}66$) & IN-V2 ($n{=}294$) & Pooled$^\ddagger$\\
\midrule
\multicolumn{7}{@{}l}{\textit{Ours $-$ CLIP ens.}}\\
CLIP B/16 & OpenAI WIT & $+0.6$\,{\scriptsize$[-1.3,\,+2.8]$} & $+1.9$\,{\scriptsize$[-0.1,\,+4.0]$} & $\mathbf{+2.3}$\,{\scriptsize$[+0.2,\,+4.6]$} & $\mathbf{+2.2}$\,{\scriptsize$[+0.7,\,+3.6]$} & $\mathbf{+1.8}$\,{\scriptsize$[+0.9,\,+2.7]$}\\
OpenCLIP B/16 & LAION-2B & $-0.7$\,{\scriptsize$[-2.8,\,+1.4]$} & $+0.7$\,{\scriptsize$[-1.4,\,+2.9]$} & $-0.8$\,{\scriptsize$[-4.2,\,+2.9]$} & $\mathbf{+1.8}$\,{\scriptsize$[+0.4,\,+3.4]$} & $+0.7$\,{\scriptsize$[-0.3,\,+1.7]$}\\
SigLIP B/16 & WebLI & $+0.5$\,{\scriptsize$[-0.8,\,+1.9]$} & $+1.0$\,{\scriptsize$[-0.7,\,+2.8]$} & $+2.7$\,{\scriptsize$[-0.5,\,+6.0]$} & $\mathbf{+1.9}$\,{\scriptsize$[+0.5,\,+3.3]$} & $\mathbf{+1.3}$\,{\scriptsize$[+0.4,\,+2.1]$}\\
EVA02-CLIP B/16 & Merged-2B & $+0.3$\,{\scriptsize$[-1.1,\,+1.5]$} & $+1.6$\,{\scriptsize$[-0.9,\,+3.9]$} & $+0.3$\,{\scriptsize$[-1.6,\,+2.2]$} & $\mathbf{+1.4}$\,{\scriptsize$[0.0,\,+2.7]$} & $\mathbf{+0.8}$\,{\scriptsize$[0.0,\,+1.6]$}\\
CLIP L/14 & OpenAI WIT & $-1.7$\,{\scriptsize$[-3.9,\,+0.4]$} & $-1.0$\,{\scriptsize$[-3.3,\,+1.3]$} & $\mathbf{+3.7}$\,{\scriptsize$[+0.6,\,+7.0]$} & $\mathbf{+1.6}$\,{\scriptsize$[+0.4,\,+2.9]$} & $+0.7$\,{\scriptsize$[-0.2,\,+1.7]$}\\
\midrule
\multicolumn{7}{@{}l}{\textit{Gate effect (Ours $-$ soft link)}}\\
CLIP B/16 & OpenAI WIT & $\mathbf{+2.8}$\,{\scriptsize$[+0.2,\,+5.6]$} & $\mathbf{+2.4}$\,{\scriptsize$[+0.8,\,+4.3]$} & $+1.8$\,{\scriptsize$[-1.1,\,+4.7]$} & $\mathbf{+5.2}$\,{\scriptsize$[+3.9,\,+6.6]$} & $\mathbf{+3.7}$\,{\scriptsize$[+2.8,\,+4.7]$}\\
OpenCLIP B/16 & LAION-2B & $\mathbf{+3.5}$\,{\scriptsize$[+1.1,\,+6.2]$} & $\mathbf{+2.1}$\,{\scriptsize$[+0.9,\,+3.5]$} & $+0.4$\,{\scriptsize$[-0.4,\,+1.3]$} & $\mathbf{+2.2}$\,{\scriptsize$[+1.1,\,+3.4]$} & $\mathbf{+1.4}$\,{\scriptsize$[+0.8,\,+2.0]$}\\
SigLIP B/16 & WebLI & $+1.7$\,{\scriptsize$[-0.1,\,+4.1]$} & $\mathbf{+2.9}$\,{\scriptsize$[+0.9,\,+5.3]$} & $+2.0$\,{\scriptsize$[-0.5,\,+4.5]$} & $\mathbf{+2.9}$\,{\scriptsize$[+1.6,\,+4.3]$} & $\mathbf{+2.6}$\,{\scriptsize$[+1.6,\,+3.5]$}\\
EVA02-CLIP B/16 & Merged-2B & $\mathbf{+3.6}$\,{\scriptsize$[+0.7,\,+7.5]$} & $\mathbf{+1.6}$\,{\scriptsize$[+0.2,\,+3.2]$} & $+2.3$\,{\scriptsize$[-0.8,\,+5.1]$} & $\mathbf{+3.6}$\,{\scriptsize$[+2.2,\,+5.1]$} & $\mathbf{+2.7}$\,{\scriptsize$[+1.8,\,+3.6]$}\\
CLIP L/14 & OpenAI WIT & $+0.9$\,{\scriptsize$[-1.8,\,+3.6]$} & $\mathbf{+2.4}$\,{\scriptsize$[+1.2,\,+3.6]$} & $-1.3$\,{\scriptsize$[-5.0,\,+2.0]$} & $\mathbf{+3.3}$\,{\scriptsize$[+1.5,\,+5.2]$} & $\mathbf{+2.2}$\,{\scriptsize$[+1.3,\,+3.1]$}\\
\midrule
\multicolumn{7}{@{}l}{\textit{Oracle link $-$ CLIP raw}}\\
CLIP B/16 & OpenAI WIT & $\mathbf{+12.7}$\,{\scriptsize$[+6.1,\,+20.4]$} & $\mathbf{+13.1}$\,{\scriptsize$[+7.5,\,+19.1]$} & $\mathbf{+17.1}$\,{\scriptsize$[+11.6,\,+22.8]$} & $\mathbf{+14.9}$\,{\scriptsize$[+10.9,\,+19.0]$} & $\mathbf{+14.7}$\,{\scriptsize$[+12.1,\,+17.3]$}\\
OpenCLIP B/16 & LAION-2B & $\mathbf{+13.2}$\,{\scriptsize$[+8.1,\,+18.7]$} & $\mathbf{+17.2}$\,{\scriptsize$[+11.3,\,+23.4]$} & $\mathbf{+17.1}$\,{\scriptsize$[+11.6,\,+23.0]$} & $\mathbf{+14.2}$\,{\scriptsize$[+10.3,\,+18.1]$} & $\mathbf{+15.0}$\,{\scriptsize$[+12.5,\,+17.5]$}\\
SigLIP B/16 & WebLI & $\mathbf{+6.7}$\,{\scriptsize$[+3.3,\,+10.8]$} & $\mathbf{+10.5}$\,{\scriptsize$[+6.3,\,+15.1]$} & $\mathbf{+19.1}$\,{\scriptsize$[+14.7,\,+23.8]$} & $\mathbf{+12.7}$\,{\scriptsize$[+9.2,\,+16.1]$} & $\mathbf{+11.8}$\,{\scriptsize$[+9.9,\,+13.8]$}\\
EVA02-CLIP B/16 & Merged-2B & $\mathbf{+13.6}$\,{\scriptsize$[+7.3,\,+20.8]$} & $\mathbf{+16.8}$\,{\scriptsize$[+10.8,\,+23.3]$} & $\mathbf{+20.2}$\,{\scriptsize$[+14.3,\,+26.6]$} & $\mathbf{+14.4}$\,{\scriptsize$[+10.4,\,+18.5]$} & $\mathbf{+15.9}$\,{\scriptsize$[+13.2,\,+18.6]$}\\
CLIP L/14 & OpenAI WIT & $\mathbf{+13.5}$\,{\scriptsize$[+7.0,\,+20.9]$} & $\mathbf{+9.5}$\,{\scriptsize$[+4.6,\,+15.0]$} & $\mathbf{+20.6}$\,{\scriptsize$[+14.5,\,+27.4]$} & $\mathbf{+14.6}$\,{\scriptsize$[+10.6,\,+18.6]$} & $\mathbf{+14.2}$\,{\scriptsize$[+11.5,\,+16.8]$}\\
\bottomrule
\end{tabular}\end{adjustbox}
\caption{\textbf{Five dual-encoder backbones, four test sets} (top-1 points, seed 0, inductive). Every cell is a paired difference with a 95\% class-level bootstrap interval; bold: the interval excludes zero. The pipeline, splits and vocabulary draws are the same in every row, and all hyper-parameters are re-selected per backbone with the rules of Sec.~\ref{sec:method}. $^\ddagger$Inverse-variance pooling over the four sets, as in Tab.~\ref{tab:diff}; approximate, because the sets share ImageNet classes. The CLIP B/16 rows reproduce Tab.~\ref{tab:diff} exactly.}
\label{tab:bb}
\end{table*}

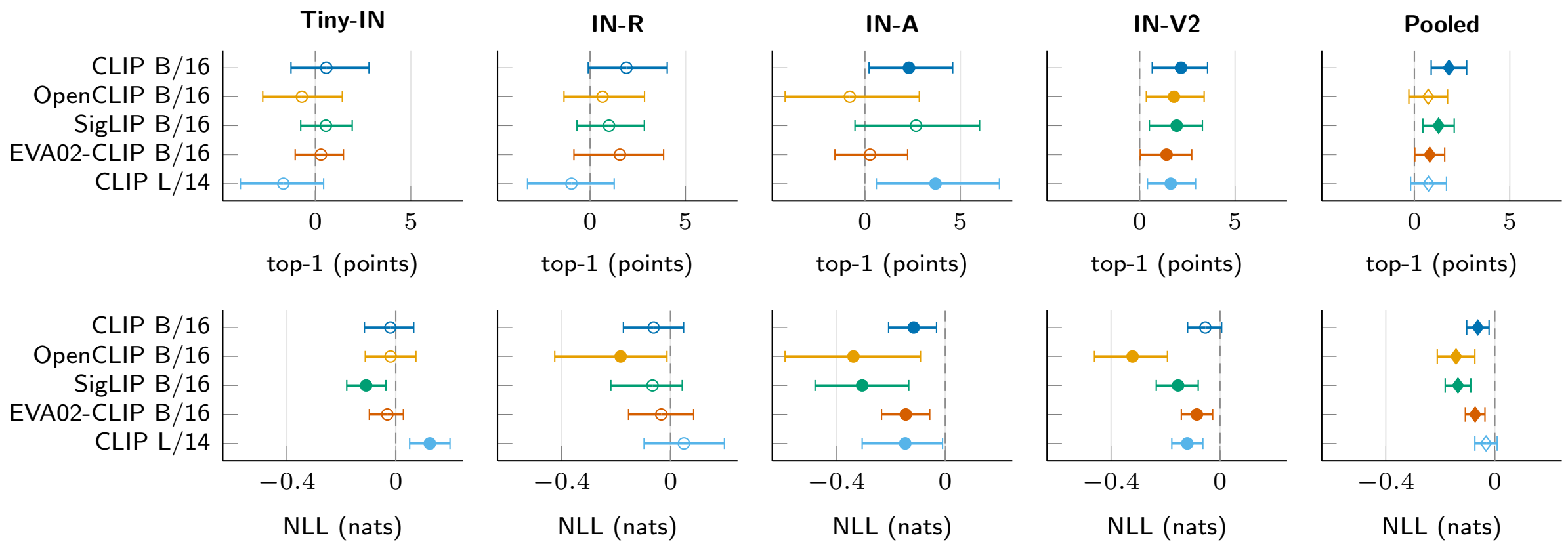
\begin{figure*}[!ht]
\centering
\begin{tikzpicture}[font=\scriptsize\sffamily]
\begin{groupplot}[group style={group size=5 by 2, horizontal sep=3.5mm, vertical sep=11mm, yticklabels at=edge left},
  width=0.225\textwidth, height=31mm, ymin=0.4, ymax=5.6, ytick={1,2,3,4,5},
  yticklabels={CLIP L/14,EVA02-CLIP B/16,SigLIP B/16,OpenCLIP B/16,CLIP B/16}, xmajorgrids, grid style={black!10}, axis lines*=left, clip=false]
\nextgroupplot[xmin=-4.854, xmax=7.724, xlabel={top-1 (points)}, xtick={-5,0,5,10,15}, title={Tiny-IN}, title style={font=\scriptsize\sffamily\bfseries, yshift=-1.5mm}]
\draw[black!45, densely dashed] (axis cs:0,0.4) -- (axis cs:0,5.6);
\addplot[cBlue, mark=o, mark size=1.6pt, mark options={fill=cBlue!0}, only marks, error bars/.cd, x dir=both, x explicit, error bar style={line width=0.7pt}] coordinates {(0.570,5) += (2.240,0) -= (1.850,0)};
\addplot[cOrange, mark=o, mark size=1.6pt, mark options={fill=cOrange!0}, only marks, error bars/.cd, x dir=both, x explicit, error bar style={line width=0.7pt}] coordinates {(-0.710,4) += (2.120,0) -= (2.050,0)};
\addplot[cGreen, mark=o, mark size=1.6pt, mark options={fill=cGreen!0}, only marks, error bars/.cd, x dir=both, x explicit, error bar style={line width=0.7pt}] coordinates {(0.550,3) += (1.380,0) -= (1.320,0)};
\addplot[cVerm, mark=o, mark size=1.6pt, mark options={fill=cVerm!0}, only marks, error bars/.cd, x dir=both, x explicit, error bar style={line width=0.7pt}] coordinates {(0.290,2) += (1.180,0) -= (1.350,0)};
\addplot[cSky, mark=o, mark size=1.6pt, mark options={fill=cSky!0}, only marks, error bars/.cd, x dir=both, x explicit, error bar style={line width=0.7pt}] coordinates {(-1.680,1) += (2.110,0) -= (2.250,0)};
\nextgroupplot[xmin=-4.854, xmax=7.724, xlabel={top-1 (points)}, xtick={-5,0,5,10,15}, title={IN-R}, title style={font=\scriptsize\sffamily\bfseries, yshift=-1.5mm}]
\draw[black!45, densely dashed] (axis cs:0,0.4) -- (axis cs:0,5.6);
\addplot[cBlue, mark=o, mark size=1.6pt, mark options={fill=cBlue!0}, only marks, error bars/.cd, x dir=both, x explicit, error bar style={line width=0.7pt}] coordinates {(1.900,5) += (2.140,0) -= (2.000,0)};
\addplot[cOrange, mark=o, mark size=1.6pt, mark options={fill=cOrange!0}, only marks, error bars/.cd, x dir=both, x explicit, error bar style={line width=0.7pt}] coordinates {(0.650,4) += (2.200,0) -= (2.020,0)};
\addplot[cGreen, mark=o, mark size=1.6pt, mark options={fill=cGreen!0}, only marks, error bars/.cd, x dir=both, x explicit, error bar style={line width=0.7pt}] coordinates {(0.990,3) += (1.850,0) -= (1.680,0)};
\addplot[cVerm, mark=o, mark size=1.6pt, mark options={fill=cVerm!0}, only marks, error bars/.cd, x dir=both, x explicit, error bar style={line width=0.7pt}] coordinates {(1.560,2) += (2.290,0) -= (2.410,0)};
\addplot[cSky, mark=o, mark size=1.6pt, mark options={fill=cSky!0}, only marks, error bars/.cd, x dir=both, x explicit, error bar style={line width=0.7pt}] coordinates {(-0.980,1) += (2.240,0) -= (2.300,0)};
\nextgroupplot[xmin=-4.854, xmax=7.724, xlabel={top-1 (points)}, xtick={-5,0,5,10,15}, title={IN-A}, title style={font=\scriptsize\sffamily\bfseries, yshift=-1.5mm}]
\draw[black!45, densely dashed] (axis cs:0,0.4) -- (axis cs:0,5.6);
\addplot[cBlue, mark=*, mark size=1.6pt, mark options={fill=cBlue}, only marks, error bars/.cd, x dir=both, x explicit, error bar style={line width=0.7pt}] coordinates {(2.310,5) += (2.290,0) -= (2.090,0)};
\addplot[cOrange, mark=o, mark size=1.6pt, mark options={fill=cOrange!0}, only marks, error bars/.cd, x dir=both, x explicit, error bar style={line width=0.7pt}] coordinates {(-0.790,4) += (3.640,0) -= (3.390,0)};
\addplot[cGreen, mark=o, mark size=1.6pt, mark options={fill=cGreen!0}, only marks, error bars/.cd, x dir=both, x explicit, error bar style={line width=0.7pt}] coordinates {(2.680,3) += (3.330,0) -= (3.200,0)};
\addplot[cVerm, mark=o, mark size=1.6pt, mark options={fill=cVerm!0}, only marks, error bars/.cd, x dir=both, x explicit, error bar style={line width=0.7pt}] coordinates {(0.270,2) += (1.970,0) -= (1.840,0)};
\addplot[cSky, mark=*, mark size=1.6pt, mark options={fill=cSky}, only marks, error bars/.cd, x dir=both, x explicit, error bar style={line width=0.7pt}] coordinates {(3.700,1) += (3.350,0) -= (3.100,0)};
\nextgroupplot[xmin=-4.854, xmax=7.724, xlabel={top-1 (points)}, xtick={-5,0,5,10,15}, title={IN-V2}, title style={font=\scriptsize\sffamily\bfseries, yshift=-1.5mm}]
\draw[black!45, densely dashed] (axis cs:0,0.4) -- (axis cs:0,5.6);
\addplot[cBlue, mark=*, mark size=1.6pt, mark options={fill=cBlue}, only marks, error bars/.cd, x dir=both, x explicit, error bar style={line width=0.7pt}] coordinates {(2.170,5) += (1.390,0) -= (1.510,0)};
\addplot[cOrange, mark=*, mark size=1.6pt, mark options={fill=cOrange}, only marks, error bars/.cd, x dir=both, x explicit, error bar style={line width=0.7pt}] coordinates {(1.800,4) += (1.580,0) -= (1.450,0)};
\addplot[cGreen, mark=*, mark size=1.6pt, mark options={fill=cGreen}, only marks, error bars/.cd, x dir=both, x explicit, error bar style={line width=0.7pt}] coordinates {(1.940,3) += (1.350,0) -= (1.430,0)};
\addplot[cVerm, mark=*, mark size=1.6pt, mark options={fill=cVerm}, only marks, error bars/.cd, x dir=both, x explicit, error bar style={line width=0.7pt}] coordinates {(1.410,2) += (1.320,0) -= (1.380,0)};
\addplot[cSky, mark=*, mark size=1.6pt, mark options={fill=cSky}, only marks, error bars/.cd, x dir=both, x explicit, error bar style={line width=0.7pt}] coordinates {(1.630,1) += (1.300,0) -= (1.220,0)};
\nextgroupplot[xmin=-4.854, xmax=7.724, xlabel={top-1 (points)}, xtick={-5,0,5,10,15}, title={Pooled}, title style={font=\scriptsize\sffamily\bfseries, yshift=-1.5mm}]
\draw[black!45, densely dashed] (axis cs:0,0.4) -- (axis cs:0,5.6);
\addplot[cBlue, mark=diamond*, mark size=2.1pt, mark options={fill=cBlue}, only marks, error bars/.cd, x dir=both, x explicit, error bar style={line width=0.7pt}] coordinates {(1.810,5) += (0.930,0) -= (0.930,0)};
\addplot[cOrange, mark=diamond, mark size=2.1pt, mark options={fill=cOrange!0}, only marks, error bars/.cd, x dir=both, x explicit, error bar style={line width=0.7pt}] coordinates {(0.724,4) += (1.015,0) -= (1.015,0)};
\addplot[cGreen, mark=diamond*, mark size=2.1pt, mark options={fill=cGreen}, only marks, error bars/.cd, x dir=both, x explicit, error bar style={line width=0.7pt}] coordinates {(1.266,3) += (0.822,0) -= (0.822,0)};
\addplot[cVerm, mark=diamond*, mark size=2.1pt, mark options={fill=cVerm}, only marks, error bars/.cd, x dir=both, x explicit, error bar style={line width=0.7pt}] coordinates {(0.805,2) += (0.783,0) -= (0.783,0)};
\addplot[cSky, mark=diamond, mark size=2.1pt, mark options={fill=cSky!0}, only marks, error bars/.cd, x dir=both, x explicit, error bar style={line width=0.7pt}] coordinates {(0.742,1) += (0.941,0) -= (0.941,0)};
\nextgroupplot[xmin=-0.635, xmax=0.246, xlabel={NLL (nats)}, xtick={-0.4,0}, x tick label style={/pgf/number format/fixed}]
\draw[black!45, densely dashed] (axis cs:0,0.4) -- (axis cs:0,5.6);
\addplot[cBlue, mark=o, mark size=1.6pt, mark options={fill=cBlue!0}, only marks, error bars/.cd, x dir=both, x explicit, error bar style={line width=0.7pt}] coordinates {(-0.020,5) += (0.086,0) -= (0.095,0)};
\addplot[cOrange, mark=o, mark size=1.6pt, mark options={fill=cOrange!0}, only marks, error bars/.cd, x dir=both, x explicit, error bar style={line width=0.7pt}] coordinates {(-0.019,4) += (0.093,0) -= (0.093,0)};
\addplot[cGreen, mark=*, mark size=1.6pt, mark options={fill=cGreen}, only marks, error bars/.cd, x dir=both, x explicit, error bar style={line width=0.7pt}] coordinates {(-0.109,3) += (0.073,0) -= (0.071,0)};
\addplot[cVerm, mark=o, mark size=1.6pt, mark options={fill=cVerm!0}, only marks, error bars/.cd, x dir=both, x explicit, error bar style={line width=0.7pt}] coordinates {(-0.031,2) += (0.059,0) -= (0.066,0)};
\addplot[cSky, mark=*, mark size=1.6pt, mark options={fill=cSky}, only marks, error bars/.cd, x dir=both, x explicit, error bar style={line width=0.7pt}] coordinates {(0.125,1) += (0.074,0) -= (0.074,0)};
\nextgroupplot[xmin=-0.635, xmax=0.246, xlabel={NLL (nats)}, xtick={-0.4,0}, x tick label style={/pgf/number format/fixed}]
\draw[black!45, densely dashed] (axis cs:0,0.4) -- (axis cs:0,5.6);
\addplot[cBlue, mark=o, mark size=1.6pt, mark options={fill=cBlue!0}, only marks, error bars/.cd, x dir=both, x explicit, error bar style={line width=0.7pt}] coordinates {(-0.062,5) += (0.110,0) -= (0.111,0)};
\addplot[cOrange, mark=*, mark size=1.6pt, mark options={fill=cOrange}, only marks, error bars/.cd, x dir=both, x explicit, error bar style={line width=0.7pt}] coordinates {(-0.183,4) += (0.170,0) -= (0.242,0)};
\addplot[cGreen, mark=o, mark size=1.6pt, mark options={fill=cGreen!0}, only marks, error bars/.cd, x dir=both, x explicit, error bar style={line width=0.7pt}] coordinates {(-0.066,3) += (0.109,0) -= (0.153,0)};
\addplot[cVerm, mark=o, mark size=1.6pt, mark options={fill=cVerm!0}, only marks, error bars/.cd, x dir=both, x explicit, error bar style={line width=0.7pt}] coordinates {(-0.034,2) += (0.119,0) -= (0.120,0)};
\addplot[cSky, mark=o, mark size=1.6pt, mark options={fill=cSky!0}, only marks, error bars/.cd, x dir=both, x explicit, error bar style={line width=0.7pt}] coordinates {(0.049,1) += (0.149,0) -= (0.146,0)};
\nextgroupplot[xmin=-0.635, xmax=0.246, xlabel={NLL (nats)}, xtick={-0.4,0}, x tick label style={/pgf/number format/fixed}]
\draw[black!45, densely dashed] (axis cs:0,0.4) -- (axis cs:0,5.6);
\addplot[cBlue, mark=*, mark size=1.6pt, mark options={fill=cBlue}, only marks, error bars/.cd, x dir=both, x explicit, error bar style={line width=0.7pt}] coordinates {(-0.116,5) += (0.084,0) -= (0.092,0)};
\addplot[cOrange, mark=*, mark size=1.6pt, mark options={fill=cOrange}, only marks, error bars/.cd, x dir=both, x explicit, error bar style={line width=0.7pt}] coordinates {(-0.337,4) += (0.246,0) -= (0.251,0)};
\addplot[cGreen, mark=*, mark size=1.6pt, mark options={fill=cGreen}, only marks, error bars/.cd, x dir=both, x explicit, error bar style={line width=0.7pt}] coordinates {(-0.305,3) += (0.171,0) -= (0.173,0)};
\addplot[cVerm, mark=*, mark size=1.6pt, mark options={fill=cVerm}, only marks, error bars/.cd, x dir=both, x explicit, error bar style={line width=0.7pt}] coordinates {(-0.145,2) += (0.088,0) -= (0.089,0)};
\addplot[cSky, mark=*, mark size=1.6pt, mark options={fill=cSky}, only marks, error bars/.cd, x dir=both, x explicit, error bar style={line width=0.7pt}] coordinates {(-0.147,1) += (0.137,0) -= (0.158,0)};
\nextgroupplot[xmin=-0.635, xmax=0.246, xlabel={NLL (nats)}, xtick={-0.4,0}, x tick label style={/pgf/number format/fixed}]
\draw[black!45, densely dashed] (axis cs:0,0.4) -- (axis cs:0,5.6);
\addplot[cBlue, mark=o, mark size=1.6pt, mark options={fill=cBlue!0}, only marks, error bars/.cd, x dir=both, x explicit, error bar style={line width=0.7pt}] coordinates {(-0.054,5) += (0.060,0) -= (0.065,0)};
\addplot[cOrange, mark=*, mark size=1.6pt, mark options={fill=cOrange}, only marks, error bars/.cd, x dir=both, x explicit, error bar style={line width=0.7pt}] coordinates {(-0.321,4) += (0.128,0) -= (0.140,0)};
\addplot[cGreen, mark=*, mark size=1.6pt, mark options={fill=cGreen}, only marks, error bars/.cd, x dir=both, x explicit, error bar style={line width=0.7pt}] coordinates {(-0.154,3) += (0.074,0) -= (0.080,0)};
\addplot[cVerm, mark=*, mark size=1.6pt, mark options={fill=cVerm}, only marks, error bars/.cd, x dir=both, x explicit, error bar style={line width=0.7pt}] coordinates {(-0.085,2) += (0.058,0) -= (0.057,0)};
\addplot[cSky, mark=*, mark size=1.6pt, mark options={fill=cSky}, only marks, error bars/.cd, x dir=both, x explicit, error bar style={line width=0.7pt}] coordinates {(-0.120,1) += (0.057,0) -= (0.057,0)};
\nextgroupplot[xmin=-0.635, xmax=0.246, xlabel={NLL (nats)}, xtick={-0.4,0}, x tick label style={/pgf/number format/fixed}]
\draw[black!45, densely dashed] (axis cs:0,0.4) -- (axis cs:0,5.6);
\addplot[cBlue, mark=diamond*, mark size=2.1pt, mark options={fill=cBlue}, only marks, error bars/.cd, x dir=both, x explicit, error bar style={line width=0.7pt}] coordinates {(-0.062,5) += (0.041,0) -= (0.041,0)};
\addplot[cOrange, mark=diamond*, mark size=2.1pt, mark options={fill=cOrange}, only marks, error bars/.cd, x dir=both, x explicit, error bar style={line width=0.7pt}] coordinates {(-0.142,4) += (0.069,0) -= (0.069,0)};
\addplot[cGreen, mark=diamond*, mark size=2.1pt, mark options={fill=cGreen}, only marks, error bars/.cd, x dir=both, x explicit, error bar style={line width=0.7pt}] coordinates {(-0.135,3) += (0.047,0) -= (0.047,0)};
\addplot[cVerm, mark=diamond*, mark size=2.1pt, mark options={fill=cVerm}, only marks, error bars/.cd, x dir=both, x explicit, error bar style={line width=0.7pt}] coordinates {(-0.072,2) += (0.036,0) -= (0.036,0)};
\addplot[cSky, mark=diamond, mark size=2.1pt, mark options={fill=cSky!0}, only marks, error bars/.cd, x dir=both, x explicit, error bar style={line width=0.7pt}] coordinates {(-0.032,1) += (0.041,0) -= (0.041,0)};
\end{groupplot}
\end{tikzpicture}
\caption{\textbf{Ours $-$ CLIP ensemble for every backbone and test set.} Top: top-1 accuracy (points, higher is better). Bottom: NLL (nats, lower is better). Points are paired differences with 95\% class-level bootstrap intervals (seed 0); diamonds are the approximate inverse-variance pool over the four sets. Filled markers: the interval excludes zero. Dashed line: no difference.}
\label{fig:bb}
\end{figure*}

\paragraph{Does it carry over to other backbones?} We rerun the inductive and transductive comparisons with five dual-encoder models on all four test sets (Tab.~\ref{tab:bb}, Fig.~\ref{fig:bb}; accuracies and the extensions are in supplementary Tabs.~\ref{tab:bbacc} and~\ref{tab:bbext}). The CLIP B/16 runs are an independent rerun of the main experiment, and all 1312 values we compared agree with it to four decimals.
\begin{itemize}
\item \emph{The shift and the headroom are general.} Renaming costs between 8.2 (SigLIP, Tiny-IN) and 22.3 points (CLIP L/14, IN-A). The oracle link beats raw CLIP significantly in all 20 backbone--set pairs, by $+6.7$ to $+20.6$ points. Link accuracy does not follow recognition accuracy: EVA02-CLIP has the highest development-name accuracy on Tiny-IN (86.8\%) but links only 32\% of its test-class names correctly, while SigLIP links best on every set.
\item \emph{Ours is ahead of the ensemble on every backbone after pooling, usually by less than on CLIP B/16.} Ours $-$ CLIP ens.\ is positive in 16 of 20 pairs, significant in 7, and significantly negative in none. On IN-V2, the largest set, it is significant for all five backbones (for EVA02-CLIP only just: the lower bound is $+0.03$ points), from $+1.4$ to $+2.2$ points. Pooled over the sets it is positive for every backbone, from $+0.7$ (OpenCLIP B/16) to $+1.8$ points (CLIP B/16), and the interval excludes zero for CLIP B/16, SigLIP B/16 and EVA02-CLIP B/16. The four negative cells belong to OpenCLIP B/16 (Tiny-IN, IN-A) and CLIP L/14 (Tiny-IN, IN-R).
\item \emph{Calibration improves.} Ours has lower NLL than the ensemble in 18 of 20 pairs (11 significantly) and lower ECE in 18 (13 significantly). Pooled NLL is lower for all five backbones and significantly so for four (Fig.~\ref{fig:bb}, bottom). The one significant increase is CLIP L/14 on Tiny-IN ($\mathbf{+0.12}$\,{\scriptsize$[+0.05,\,+0.20]$} nats).
\item \emph{The gate is still what makes it work.} Without it, the soft link is below the ensemble in 17 of 20 pairs. The gate effect is positive in 19 pairs, significant in 13, and significant after pooling for every backbone ($+1.4$ to $+3.7$ points).
\item \emph{The negative results replicate.} Fused hard linking is below CLIP-memory hard linking in all 20 pairs (14 significantly, and significantly after pooling for every backbone). Transduction alone helps in 19 pairs, all significant. Transductive link re-estimation on top of it is positive in 8 pairs and negative in 12. Its pooled interval covers zero for every backbone, and its effect varies strongly across sets ($I^2=0.45$--$0.80$).
\end{itemize}

\paragraph{A second seed.} For CLIP B/16 we also ran seed 1, which draws a new class split and new held-out names (supplementary Tab.~\ref{tab:seed}). The IN-V2 gain over the ensemble replicates ($\mathbf{+2.1}$\,{\scriptsize$[+1.1,\,+3.3]$}). The IN-A gain does not ($+0.1$\,{\scriptsize$[-2.6,\,+3.2]$}), and Tiny-IN and IN-R stay non-significant. The oracle headroom and the fused hard-link deficit replicate on all four sets, and the gate effect on three of them (not IN-A). We therefore treat the IN-V2 gain as the most reliable single result and the IN-A gain in Tab.~\ref{tab:diff} as less certain.

\section{Ablations and checks}
\label{sec:ablations}

\paragraph{Pre-registered criteria.} Before this run we fixed three checks. Fused gated $-$ CLIP ens.\ $>0$ failed on every set. Transductive variant $-$ CLIP ens.\ + transduction $>0$ failed on every set. Fused top-1 $>$ memory top-1 on all held-out strings passed on every set. Our main method was the method of the previous protocol version, and the two extensions were its pre-registered candidates for improvement; since both failed, the method stays unchanged. Its own earlier criterion (Ours $-$ CLIP ens.\ on ImageNet-1k pooled over three seeds) cannot be checked yet: we have four test sets with one seed each, and ImageNet-1k validation was not run.

\paragraph{Learned text adaptor.} A residual MLP on CLIP text, trained on lexicon strings with linking, contrastive, and anchor losses, does not beat the training-free memory (supplementary Tab.~\ref{tab:adaptor}). Soft link + gate with the adaptor $-$ Ours is $+0.5$\,{\scriptsize$[-2.0,\,+2.7]$} on Tiny-IN. Adding an image cross-entropy ($\mathcal L_{\mathrm{img}}$) gives $\mathbf{+2.3}$\,{\scriptsize$[+0.6,\,+4.4]$} over CLIP ens.\ in this run, but its effect relative to the same model without $\mathcal L_{\mathrm{img}}$ is $+1.2$\,{\scriptsize$[-0.3,\,+3.2]$}. In the earlier run with the same seed it was $-0.7$ $[-1.9,0.5]$, so we do not count it as a reliable gain.

\paragraph{Other checks.} $\gamma>0$ (image-mean prototypes) was never selected. Capped column normalization of the class-to-concept coupling changed no entry by more than $6\times10^{-8}$. When hard links are wrong, the median WordNet path to the true concept is 7--12 edges on every set, so errors are rarely near-synonyms.


\section{Conclusion}
Renaming classes, with nothing else changed, costs a frozen dual encoder between 8 and 22 points across the five models and four test sets we studied, and a perfect linker would recover most of it. Vocabulary shift is therefore a property of how these models are used, not of one checkpoint. A training-free, image-grounded linker with a reliability gate recovers a small part of the loss. Over a template ensemble it gains $+1.8$ points pooled with CLIP B/16 and $+0.7$ to $+1.3$ with the other four backbones, it is never significantly worse, and it lowers pooled NLL. Our attempts to do better show where the problem lies. Linking generic lexicon strings can be improved a lot without helping, link accuracy does not grow with backbone quality, and pooling evidence over unlabeled images helps only when the images are typical of their class. All three point to the same next step: lexicon-to-class linking learned or calibrated on visually grounded concepts.

\section*{LLM Usage Statement}
The authors used LLMs for language editing, 
grammar correction, and improving the clarity of the manuscript. 
All technical content, methodology, experiments, results, and 
scientific claims were developed and verified by the authors, 
who take full responsibility for the submitted work.

\bibliographystyle{unsrt}
\bibliography{refs}

\clearpage
\onecolumn
\appendix
\section{Limitations}
Most results use one seed. The second seed exists only for CLIP B/16, and it did not reproduce the IN-A gain, so single-set significance should be read with care; the IN-V2 gain is the one that holds across seeds and backbones. All five backbones are contrastive dual encoders with a single temperature, and SigLIP is used under a softmax it was not trained with. Captioning-based and generative VLMs are not covered. The test sets have 65--294 classes and share ImageNet classes, so the pooled intervals are approximate, and we do not pool across backbones, which see the same images. Held-out WordNet lemmas include rare words, and the shift they induce may differ from real naming differences; the medical arms (UMLS vocabularies, expert-checked) remain future work. The lexicon must contain the concept for linking to help, and the gate is the only defence when it does not. Several selected hyper-parameters sit at grid edges.
\section*{Supplementary material}
\setcounter{table}{5}

\begin{table*}[h!]
\centering\small
\setlength{\tabcolsep}{3pt}
\begin{adjustbox}{max width=\textwidth}\begin{tabular}{@{}lcccccccccccc@{}}
\toprule
 & \multicolumn{6}{c}{Tiny-IN} & \multicolumn{6}{c}{IN-R}\\
\cmidrule(lr){2-7}\cmidrule(lr){8-13}
Method & Acc & Worst & All & Cons & NLL & ECE & Acc & Worst & All & Cons & NLL & ECE\\
\midrule
CLIP raw & 60.8 & 59.9 & 57.6 & 89.6 & 1.544 & 4.2 & 65.1 & 63.9 & 62.4 & 93.5 & 1.434 & 4.5\\
CLIP ens. & 61.1 & 60.2 & 57.8 & 90.1 & 1.524 & 3.9 & 66.1 & 64.8 & 63.2 & 93.7 & 1.386 & 4.4\\
Soft link & 58.9 & 58.2 & 56.6 & 91.9 & 1.656 & 5.7 & 65.6 & 64.8 & 62.4 & 92.7 & 1.558 & 3.2\\
\textbf{Ours} & 61.6 & 60.7 & 58.8 & 91.5 & 1.504 & 3.4 & 68.0 & 67.2 & 65.1 & 93.5 & 1.371 & 2.4\\
Ours + fused & 61.1 & 60.4 & 58.8 & 92.2 & 1.531 & 4.7 & 65.8 & 65.3 & 63.2 & 93.5 & 1.395 & 1.7\\
CLIP ens.\ + T & 63.8 & 62.4 & 59.7 & 89.4 & 2.029 & 19.7 & 67.8 & 66.8 & 65.4 & 94.1 & 2.020 & 19.0\\
Link EM + cent. & 67.2 & 67.0 & 64.8 & 92.4 & 1.716 & 16.1 & 63.6 & 63.1 & 59.5 & 88.5 & 2.496 & 23.5\\
Oracle link & 73.5 & 73.4 & 73.1 & 99.4 & 0.983 & 4.5 & 78.2 & 77.8 & 77.7 & 98.6 & 0.797 & 1.4\\
\midrule
 & \multicolumn{6}{c}{IN-A} & \multicolumn{6}{c}{IN-V2}\\
\cmidrule(lr){2-7}\cmidrule(lr){8-13}
Method & Acc & Worst & All & Cons & NLL & ECE & Acc & Worst & All & Cons & NLL & ECE\\
\midrule
CLIP raw & 43.8 & 43.1 & 42.0 & 94.3 & 2.256 & 14.2 & 55.9 & 55.1 & 52.0 & 89.0 & 2.081 & 5.1\\
CLIP ens. & 44.4 & 44.0 & 42.6 & 94.6 & 2.222 & 14.3 & 56.7 & 56.0 & 53.0 & 89.5 & 2.052 & 5.4\\
Soft link & 45.0 & 44.5 & 43.3 & 96.0 & 2.215 & 9.3 & 53.6 & 52.9 & 49.9 & 89.5 & 2.255 & 3.6\\
\textbf{Ours} & 46.8 & 46.3 & 45.1 & 95.1 & 2.113 & 10.7 & 58.9 & 58.4 & 55.2 & 90.2 & 1.998 & 3.5\\
Ours + fused & 46.7 & 46.1 & 45.0 & 96.2 & 2.108 & 7.3 & 58.0 & 57.3 & 54.4 & 90.0 & 1.939 & 5.3\\
CLIP ens.\ + T & 49.1 & 48.7 & 47.2 & 93.5 & 2.315 & 25.2 & 58.7 & 57.7 & 53.9 & 86.2 & 4.544 & 36.6\\
Link EM + cent. & 45.4 & 45.2 & 43.4 & 94.9 & 3.775 & 35.8 & 60.9 & 60.1 & 56.4 & 88.1 & 3.025 & 27.3\\
Oracle link & 60.9 & 60.3 & 59.8 & 98.5 & 1.455 & 5.2 & 70.7 & 70.1 & 68.9 & 96.3 & 1.237 & 2.7\\
\bottomrule
\end{tabular}\end{adjustbox}
\caption{\textbf{All metrics} for the main arms on every test set (one seed). Acc: mean top-1 over three vocabulary draws. Worst: worst draw. All: images correct under all draws. Cons: pairwise prediction consistency. ECE in \%, 15 bins. NLL in nats. ``+ T'': transductive.}
\label{tab:full}
\end{table*}

\begin{table*}[h!]
\centering\small
\setlength{\tabcolsep}{5pt}
\begin{adjustbox}{max width=\textwidth}\begin{tabular}{@{}llcccc@{}}
\toprule
Comparison & Metric & Tiny-IN & IN-R & IN-A & IN-V2\\
\midrule
Ours $-$ CLIP ens. & All $\uparrow$ & $+1.1$\,{\scriptsize$[-0.9,\,+3.3]$} & $+1.8$\,{\scriptsize$[-0.2,\,+3.9]$} & $\mathbf{+2.5}$\,{\scriptsize$[+0.4,\,+4.8]$} & $\mathbf{+2.2}$\,{\scriptsize$[+0.6,\,+3.6]$}\\
 & Cons $\uparrow$ & $\mathbf{+1.3}$\,{\scriptsize$[+0.4,\,+2.5]$} & $+0.2$\,{\scriptsize$[-0.8,\,+1.0]$} & $+0.4$\,{\scriptsize$[-0.4,\,+1.1]$} & $+0.7$\,{\scriptsize$[-0.3,\,+1.5]$}\\
 & NLL $\downarrow$ & $-0.020$\,{\scriptsize$[-0.115,\,+0.066]$} & $-0.062$\,{\scriptsize$[-0.173,\,+0.048]$} & $\mathbf{-0.116}$\,{\scriptsize$[-0.208,\,-0.033]$} & $-0.054$\,{\scriptsize$[-0.119,\,+0.006]$}\\
 & ECE $\downarrow$ & $-0.6$\,{\scriptsize$[-5.1,\,+5.7]$} & $-2.0$\,{\scriptsize$[-3.4,\,+1.4]$} & $\mathbf{-3.6}$\,{\scriptsize$[-5.0,\,-2.1]$} & $-1.9$\,{\scriptsize$[-4.0,\,+2.2]$}\\
\addlinespace[2pt]
Ours + fused $-$ CLIP ens. & All $\uparrow$ & $+1.1$\,{\scriptsize$[-2.5,\,+4.8]$} & $-0.1$\,{\scriptsize$[-2.3,\,+2.1]$} & $+2.4$\,{\scriptsize$[-0.5,\,+5.3]$} & $+1.3$\,{\scriptsize$[-0.7,\,+3.4]$}\\
 & Cons $\uparrow$ & $\mathbf{+2.0}$\,{\scriptsize$[+0.8,\,+3.4]$} & $-0.7$\,{\scriptsize$[-2.0,\,+0.7]$} & $\mathbf{+1.5}$\,{\scriptsize$[+0.2,\,+3.0]$} & $+0.5$\,{\scriptsize$[-0.6,\,+1.5]$}\\
 & NLL $\downarrow$ & $+0.007$\,{\scriptsize$[-0.114,\,+0.116]$} & $-0.001$\,{\scriptsize$[-0.160,\,+0.153]$} & $-0.069$\,{\scriptsize$[-0.193,\,+0.036]$} & $\mathbf{-0.113}$\,{\scriptsize$[-0.212,\,-0.015]$}\\
 & ECE $\downarrow$ & $+0.8$\,{\scriptsize$[-4.5,\,+6.9]$} & $-2.8$\,{\scriptsize$[-6.1,\,+4.8]$} & $\mathbf{-7.0}$\,{\scriptsize$[-9.6,\,-4.0]$} & $-0.1$\,{\scriptsize$[-3.8,\,+4.1]$}\\
\addlinespace[2pt]
Link EM + cent.\ $-$ CLIP ens.\ + T & All $\uparrow$ & $\mathbf{+5.1}$\,{\scriptsize$[+1.2,\,+10.2]$} & $\mathbf{-5.9}$\,{\scriptsize$[-10.3,\,-2.2]$} & $\mathbf{-3.8}$\,{\scriptsize$[-7.5,\,0.0]$} & $+2.5$\,{\scriptsize$[-0.1,\,+5.2]$}\\
 & Cons $\uparrow$ & $\mathbf{+3.1}$\,{\scriptsize$[+0.3,\,+6.0]$} & $\mathbf{-6.2}$\,{\scriptsize$[-8.7,\,-4.0]$} & $+1.3$\,{\scriptsize$[-0.9,\,+3.8]$} & $\mathbf{+1.9}$\,{\scriptsize$[+0.4,\,+3.5]$}\\
 & NLL $\downarrow$ & $\mathbf{-0.314}$\,{\scriptsize$[-0.677,\,-0.032]$} & $\mathbf{+0.431}$\,{\scriptsize$[+0.053,\,+0.863]$} & $\mathbf{+1.805}$\,{\scriptsize$[+1.056,\,+2.541]$} & $\mathbf{-1.519}$\,{\scriptsize$[-1.852,\,-1.202]$}\\
 & ECE $\downarrow$ & $\mathbf{-3.6}$\,{\scriptsize$[-7.6,\,-0.6]$} & $\mathbf{+4.5}$\,{\scriptsize$[+0.7,\,+10.1]$} & $\mathbf{+10.6}$\,{\scriptsize$[+4.9,\,+15.5]$} & $\mathbf{-9.3}$\,{\scriptsize$[-11.7,\,-6.9]$}\\
\bottomrule
\end{tabular}\end{adjustbox}
\caption{\textbf{Robustness and calibration differences} with 95\% class-level bootstrap intervals (1000 resamples; ECE on draws pooled over vocabularies). All/Cons/ECE in points, NLL in nats. Bold: interval excludes zero.}
\label{tab:extra}
\end{table*}

\begin{table*}[h!]
\centering\small
\setlength{\tabcolsep}{5pt}
\begin{adjustbox}{max width=\textwidth}\begin{tabular}{@{}llcccc@{}}
\toprule
Variant & Parameter (selected on) & Tiny-IN & IN-R & IN-A & IN-V2\\
\midrule
Ours & $M$, $\tau_\ell$ (val.\ classes, acc.) & 160, 0.02 & 3, 0.005 & 40, 0.01 & 10, 0.05\\
 & $\varepsilon$, $\rho$ (val.\ classes, NLL) & 0.934, 0.1 & 0.918, 2 & 0.939, 0.05 & 0.924, 1\\
Fused & $\beta$, $\delta$ (3000 strings, top-1) & 0.5, 0.5 & 0.5, 0.5 & 0.5, 1 & 0.5, 0.5\\
 & $\tau_\ell$ (3000 strings, NLL) & 0.029 & 0.029 & 0.024 & 0.029\\
 & $M$ (val.\ classes, acc.) & 320 & 160 & 40 & 320\\
Link EM & $\alpha$, $\lambda$ (val.\ pool, acc.) & 3, 0.5 & 0.1, 0.5 & 0.1, 0.5 & 1, 0.25\\
CLIP ens.\ + T & $\lambda$ (val.\ pool, acc.) & 0.5 & 0.5 & 0.25 & 0.5\\
\bottomrule
\end{tabular}\end{adjustbox}
\caption{\textbf{Selected hyper-parameters} per test set. $\gamma=0$ was selected for every arm on every set. Several values sit at grid edges (see text).}
\label{tab:sel}
\end{table*}

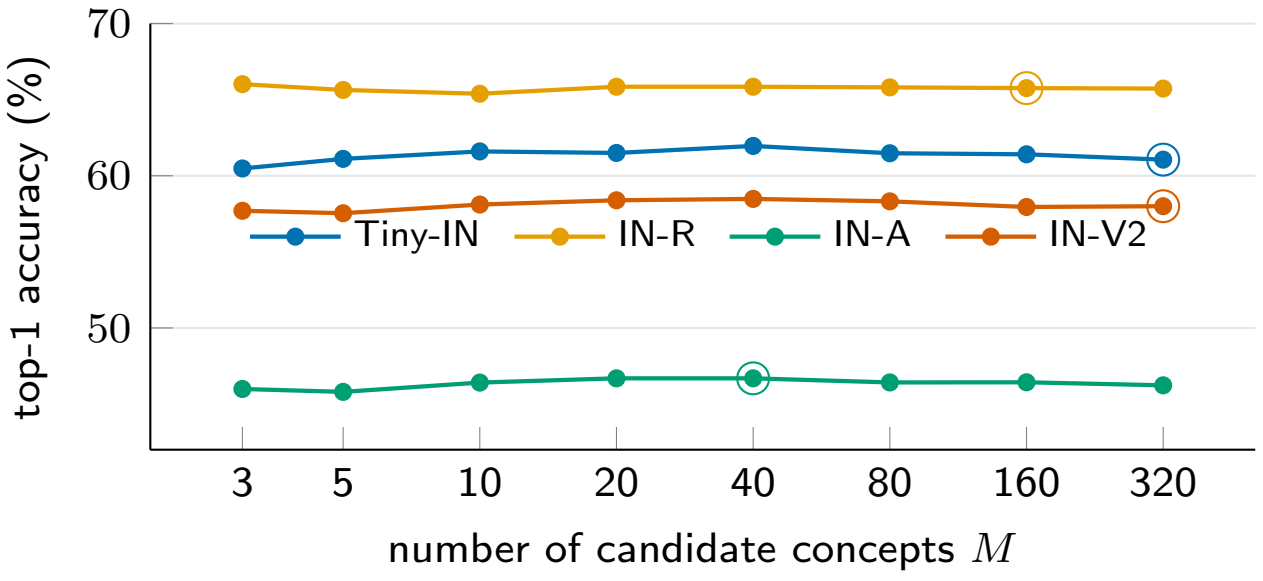
\begin{figure}[h!]
\centering
\begin{minipage}{0.5\textwidth}
\centering
\begin{tikzpicture}
\begin{axis}[font=\footnotesize\sffamily, width=\linewidth, height=44mm, xmode=log, log basis x=2, xtick={3,5,10,20,40,80,160,320},
  xticklabels={3,5,10,20,40,80,160,320}, xlabel={number of candidate concepts $M$}, ylabel={top-1 accuracy (\%)}, axis lines*=left, ymajorgrids, grid style={black!10},
  legend style={font=\scriptsize\sffamily, draw=none, fill=none, at={(0.5,0.5)}, anchor=center, legend columns=4, /tikz/every even column/.append style={column sep=4pt}}, ymin=42, ymax=70]
\addplot[cBlue, thick, mark=*, mark size=1.3pt] coordinates {(3,60.48) (5,61.11) (10,61.60) (20,61.50) (40,61.96) (80,61.48) (160,61.41) (320,61.06)}; \addlegendentry{Tiny-IN}
\addplot[cBlue, only marks, mark=o, mark size=3pt, forget plot] coordinates {(320,61.06)};
\addplot[cOrange, thick, mark=*, mark size=1.3pt] coordinates {(3,66.02) (5,65.64) (10,65.39) (20,65.85) (40,65.85) (80,65.81) (160,65.76) (320,65.73)}; \addlegendentry{IN-R}
\addplot[cOrange, only marks, mark=o, mark size=3pt, forget plot] coordinates {(160,65.76)};
\addplot[cGreen, thick, mark=*, mark size=1.3pt] coordinates {(3,45.99) (5,45.80) (10,46.41) (20,46.69) (40,46.69) (80,46.42) (160,46.43) (320,46.23)}; \addlegendentry{IN-A}
\addplot[cGreen, only marks, mark=o, mark size=3pt, forget plot] coordinates {(40,46.69)};
\addplot[cVerm, thick, mark=*, mark size=1.3pt] coordinates {(3,57.70) (5,57.54) (10,58.11) (20,58.39) (40,58.48) (80,58.32) (160,57.95) (320,58.00)}; \addlegendentry{IN-V2}
\addplot[cVerm, only marks, mark=o, mark size=3pt, forget plot] coordinates {(320,58.00)};
\end{axis}
\end{tikzpicture}
\caption{\textbf{Sensitivity to $M$} of the fused gated variant on each test set (diagnostic, test data, not used for selection; circles mark the selected $M$). Accuracy varies by at most 1.5 points over $M\in[3,320]$ once the link temperature and the gate are calibrated on text strings.}
\label{fig:msens}
\end{minipage}
\end{figure}

\begin{table}[h!]
\centering\small
\setlength{\tabcolsep}{5pt}
\begin{tabular}{@{}lcc@{}}
\toprule
Learned adaptor (Tiny-IN) & text losses & + $\mathcal L_{\mathrm{img}}$\\
\midrule
Adaptor only & 54.5 & --\\
Hard link & 56.0 & 57.7\\
Soft link & 61.1 & 63.0\\
Soft link + gate & 62.1 & 63.4\\
\midrule
Ours (training-free), for reference & 61.6 & 61.6\\
\bottomrule
\end{tabular}
\caption{\textbf{Learned text adaptor} (ablation, Tiny-IN only), top-1 \%. $\mathcal L_{\mathrm{img}}$ adds an image cross-entropy through Eq.~\eqref{eq:mix} on seen classes.}
\label{tab:adaptor}
\end{table}

\begin{table*}[h!]
\centering\small
\setlength{\tabcolsep}{4.5pt}
\begin{adjustbox}{max width=\textwidth}\begin{tabular}{@{}llcccccccc@{}}
\toprule
Backbone & Test set & Dev.\ name & Gap & CLIP raw & CLIP ens. & Soft link & \textbf{Ours} & Oracle & Link acc.\ (mem./fused)\\
\midrule
\multirow{4}{*}{CLIP B/16} & Tiny-IN & \textcolor{cGrey}{75.5} & 14.8 & 60.8 & 61.1 & 58.9 & \textbf{61.6} & \textcolor{cGrey}{73.5} & 35.9 / 34.4\\
 & IN-R & \textcolor{cGrey}{81.6} & 16.5 & 65.1 & 66.1 & 65.6 & \textbf{68.0} & \textcolor{cGrey}{78.2} & 52.9 / 43.1\\
 & IN-A & \textcolor{cGrey}{61.3} & 17.6 & 43.8 & 44.4 & 45.0 & \textbf{46.8} & \textcolor{cGrey}{60.9} & 36.9 / 35.4\\
 & IN-V2 & \textcolor{cGrey}{70.8} & 14.9 & 55.9 & 56.7 & 53.6 & \textbf{58.9} & \textcolor{cGrey}{70.7} & 36.9 / 35.6\\
\midrule
\multirow{4}{*}{OpenCLIP B/16} & Tiny-IN & \textcolor{cGrey}{77.5} & 14.3 & 63.2 & \textbf{63.9} & 59.7 & 63.2 & \textcolor{cGrey}{76.4} & 36.4 / 36.4\\
 & IN-R & \textcolor{cGrey}{84.2} & 19.0 & 65.3 & 66.2 & 64.7 & \textbf{66.8} & \textcolor{cGrey}{82.5} & 50.0 / 45.1\\
 & IN-A & \textcolor{cGrey}{50.3} & 17.0 & 33.2 & \textbf{34.4} & 33.2 & 33.6 & \textcolor{cGrey}{50.3} & 31.8 / 33.8\\
 & IN-V2 & \textcolor{cGrey}{70.0} & 15.1 & 54.9 & 55.5 & 55.1 & \textbf{57.4} & \textcolor{cGrey}{69.1} & 36.7 / 34.8\\
\midrule
\multirow{4}{*}{SigLIP B/16} & Tiny-IN & \textcolor{cGrey}{64.7} & 8.2 & 56.5 & 57.2 & 56.1 & \textbf{57.8} & \textcolor{cGrey}{63.3} & 44.6 / 39.0\\
 & IN-R & \textcolor{cGrey}{91.8} & 13.0 & 78.8 & 80.3 & 78.4 & \textbf{81.2} & \textcolor{cGrey}{89.4} & 63.7 / 56.4\\
 & IN-A & \textcolor{cGrey}{59.1} & 19.7 & 39.5 & 39.3 & 40.0 & \textbf{42.0} & \textcolor{cGrey}{58.6} & 45.0 / 38.9\\
 & IN-V2 & \textcolor{cGrey}{76.7} & 13.0 & 63.7 & 63.9 & 62.9 & \textbf{65.8} & \textcolor{cGrey}{76.4} & 46.3 / 44.3\\
\midrule
\multirow{4}{*}{EVA02-CLIP B/16} & Tiny-IN & \textcolor{cGrey}{86.8} & 14.5 & 72.3 & 72.4 & 69.0 & \textbf{72.6} & \textcolor{cGrey}{85.9} & 32.3 / 34.9\\
 & IN-R & \textcolor{cGrey}{85.1} & 19.0 & 66.1 & 66.8 & 66.7 & \textbf{68.4} & \textcolor{cGrey}{82.8} & 51.5 / 42.2\\
 & IN-A & \textcolor{cGrey}{64.8} & 20.6 & 44.2 & 44.3 & 42.3 & \textbf{44.6} & \textcolor{cGrey}{64.4} & 36.4 / 36.4\\
 & IN-V2 & \textcolor{cGrey}{75.1} & 14.8 & 60.3 & 60.2 & 58.0 & \textbf{61.6} & \textcolor{cGrey}{74.7} & 36.2 / 35.1\\
\midrule
\multirow{4}{*}{CLIP L/14} & Tiny-IN & \textcolor{cGrey}{82.7} & 14.7 & 68.0 & \textbf{68.7} & 66.1 & 67.0 & \textcolor{cGrey}{81.5} & 37.4 / 41.0\\
 & IN-R & \textcolor{cGrey}{89.7} & 11.9 & 77.7 & \textbf{77.9} & 74.6 & 77.0 & \textcolor{cGrey}{87.3} & 54.4 / 50.5\\
 & IN-A & \textcolor{cGrey}{78.2} & 22.3 & 55.9 & 55.9 & \textbf{60.8} & 59.5 & \textcolor{cGrey}{76.5} & 39.4 / 35.9\\
 & IN-V2 & \textcolor{cGrey}{76.9} & 15.2 & 61.7 & 62.7 & 61.0 & \textbf{64.3} & \textcolor{cGrey}{76.3} & 37.3 / 39.1\\
\bottomrule
\end{tabular}\end{adjustbox}
\caption{\textbf{Backbone study: top-1 accuracy (\%)} on the held-out names (seed 0, inductive). Dev.\ name: raw CLIP with development names (grey, reference). Gap: dev.\ name $-$ CLIP raw. Bold: best non-oracle method. Link acc.: fraction of test-class names whose top-1 link is the true concept, with the CLIP memory and with the fused linker, averaged over the three draws (OOL classes count as wrong).}
\label{tab:bbacc}
\end{table*}

\begin{table*}[h!]
\centering\small
\setlength{\tabcolsep}{5pt}
\begin{adjustbox}{max width=\textwidth}\begin{tabular}{@{}lcccccc@{}}
\toprule
Backbone & Tiny-IN & IN-R & IN-A & IN-V2 & Pooled & $I^2$\\
\midrule
\multicolumn{7}{@{}l}{\textit{Fused hard $-$ CLIP-memory hard link}}\\
CLIP B/16 & $\mathbf{-8.1}$\,{\scriptsize$[-14.7,\,-1.2]$} & $\mathbf{-10.2}$\,{\scriptsize$[-16.7,\,-4.5]$} & $-4.5$\,{\scriptsize$[-9.7,\,+0.2]$} & $\mathbf{-6.0}$\,{\scriptsize$[-9.2,\,-3.1]$} & $\mathbf{-6.5}$\,{\scriptsize$[-8.7,\,-4.2]$} & 0.00\\
OpenCLIP B/16 & $\mathbf{-12.8}$\,{\scriptsize$[-19.4,\,-6.7]$} & $-2.4$\,{\scriptsize$[-8.8,\,+4.0]$} & $\mathbf{-6.4}$\,{\scriptsize$[-11.5,\,-1.8]$} & $\mathbf{-7.7}$\,{\scriptsize$[-11.0,\,-4.7]$} & $\mathbf{-7.4}$\,{\scriptsize$[-9.7,\,-5.1]$} & 0.44\\
SigLIP B/16 & $\mathbf{-9.6}$\,{\scriptsize$[-15.1,\,-4.3]$} & $\mathbf{-6.3}$\,{\scriptsize$[-11.2,\,-1.7]$} & $\mathbf{-7.4}$\,{\scriptsize$[-12.6,\,-2.4]$} & $\mathbf{-3.7}$\,{\scriptsize$[-6.1,\,-1.1]$} & $\mathbf{-5.4}$\,{\scriptsize$[-7.3,\,-3.5]$} & 0.40\\
EVA02-CLIP B/16 & $-5.4$\,{\scriptsize$[-12.3,\,+1.9]$} & $\mathbf{-6.6}$\,{\scriptsize$[-13.3,\,-0.6]$} & $-4.4$\,{\scriptsize$[-10.0,\,+0.5]$} & $\mathbf{-7.1}$\,{\scriptsize$[-10.5,\,-3.7]$} & $\mathbf{-6.3}$\,{\scriptsize$[-8.7,\,-3.8]$} & 0.00\\
CLIP L/14 & $-5.9$\,{\scriptsize$[-12.7,\,+0.9]$} & $\mathbf{-7.6}$\,{\scriptsize$[-13.2,\,-2.3]$} & $\mathbf{-6.7}$\,{\scriptsize$[-11.3,\,-2.8]$} & $-1.8$\,{\scriptsize$[-4.7,\,+0.8]$} & $\mathbf{-4.1}$\,{\scriptsize$[-6.1,\,-2.1]$} & 0.50\\
\midrule
\multicolumn{7}{@{}l}{\textit{Fused gated $-$ Ours}}\\
CLIP B/16 & $-0.6$\,{\scriptsize$[-3.1,\,+1.8]$} & $-2.2$\,{\scriptsize$[-5.7,\,+0.7]$} & $-0.1$\,{\scriptsize$[-1.6,\,+1.4]$} & $-0.9$\,{\scriptsize$[-2.5,\,+0.9]$} & $-0.6$\,{\scriptsize$[-1.6,\,+0.4]$} & 0.00\\
OpenCLIP B/16 & $-0.5$\,{\scriptsize$[-3.5,\,+2.6]$} & $-0.6$\,{\scriptsize$[-3.6,\,+2.4]$} & $-0.4$\,{\scriptsize$[-4.0,\,+2.8]$} & $\mathbf{-1.8}$\,{\scriptsize$[-3.3,\,-0.1]$} & $\mathbf{-1.2}$\,{\scriptsize$[-2.4,\,0.0]$} & 0.00\\
SigLIP B/16 & $-0.7$\,{\scriptsize$[-2.3,\,+0.9]$} & $\mathbf{-3.0}$\,{\scriptsize$[-5.7,\,-1.0]$} & $\mathbf{-2.7}$\,{\scriptsize$[-5.1,\,-0.1]$} & $+0.1$\,{\scriptsize$[-0.9,\,+1.1]$} & $-0.7$\,{\scriptsize$[-1.5,\,+0.1]$} & 0.64\\
EVA02-CLIP B/16 & $+0.2$\,{\scriptsize$[-2.1,\,+2.3]$} & $-0.4$\,{\scriptsize$[-3.2,\,+2.9]$} & $-0.4$\,{\scriptsize$[-2.7,\,+1.8]$} & $-0.4$\,{\scriptsize$[-2.2,\,+1.7]$} & $-0.2$\,{\scriptsize$[-1.4,\,+0.9]$} & 0.00\\
CLIP L/14 & $\mathbf{+3.0}$\,{\scriptsize$[+0.1,\,+6.2]$} & $+0.4$\,{\scriptsize$[-2.1,\,+2.9]$} & $-0.9$\,{\scriptsize$[-3.5,\,+1.6]$} & $0.0$\,{\scriptsize$[-1.7,\,+1.8]$} & $+0.3$\,{\scriptsize$[-0.8,\,+1.5]$} & 0.25\\
\midrule
\multicolumn{7}{@{}l}{\textit{CLIP ens.\ + transd.\ $-$ CLIP ens.}}\\
CLIP B/16 & $\mathbf{+2.8}$\,{\scriptsize$[+1.5,\,+4.1]$} & $\mathbf{+1.8}$\,{\scriptsize$[+0.1,\,+3.5]$} & $\mathbf{+4.7}$\,{\scriptsize$[+2.3,\,+7.1]$} & $\mathbf{+1.9}$\,{\scriptsize$[+0.5,\,+3.4]$} & $\mathbf{+2.5}$\,{\scriptsize$[+1.7,\,+3.3]$} & 0.34\\
OpenCLIP B/16 & $\mathbf{+4.8}$\,{\scriptsize$[+2.3,\,+7.1]$} & $-0.3$\,{\scriptsize$[-2.1,\,+1.4]$} & $\mathbf{+6.2}$\,{\scriptsize$[+3.3,\,+9.0]$} & $\mathbf{+4.5}$\,{\scriptsize$[+3.3,\,+5.7]$} & $\mathbf{+3.5}$\,{\scriptsize$[+2.6,\,+4.3]$} & 0.88\\
SigLIP B/16 & $\mathbf{+3.3}$\,{\scriptsize$[+1.9,\,+4.8]$} & $\mathbf{+3.5}$\,{\scriptsize$[+1.7,\,+5.7]$} & $\mathbf{+5.9}$\,{\scriptsize$[+2.9,\,+9.0]$} & $\mathbf{+3.3}$\,{\scriptsize$[+1.9,\,+4.8]$} & $\mathbf{+3.6}$\,{\scriptsize$[+2.7,\,+4.4]$} & 0.00\\
EVA02-CLIP B/16 & $\mathbf{+3.4}$\,{\scriptsize$[+0.2,\,+7.7]$} & $\mathbf{+2.5}$\,{\scriptsize$[+1.3,\,+3.9]$} & $\mathbf{+4.7}$\,{\scriptsize$[+2.0,\,+7.7]$} & $\mathbf{+2.8}$\,{\scriptsize$[+1.4,\,+4.2]$} & $\mathbf{+2.9}$\,{\scriptsize$[+2.0,\,+3.8]$} & 0.00\\
CLIP L/14 & $\mathbf{+6.8}$\,{\scriptsize$[+4.5,\,+9.3]$} & $\mathbf{+3.5}$\,{\scriptsize$[+1.5,\,+5.5]$} & $\mathbf{+7.6}$\,{\scriptsize$[+4.1,\,+10.9]$} & $\mathbf{+3.2}$\,{\scriptsize$[+1.9,\,+4.5]$} & $\mathbf{+4.2}$\,{\scriptsize$[+3.3,\,+5.1]$} & 0.74\\
\midrule
\multicolumn{7}{@{}l}{\textit{Link EM + centroids $-$ CLIP ens.\ + transd.}}\\
CLIP B/16 & $+3.3$\,{\scriptsize$[-0.3,\,+7.7]$} & $\mathbf{-4.2}$\,{\scriptsize$[-8.4,\,-0.9]$} & $-3.7$\,{\scriptsize$[-7.2,\,0.0]$} & $+2.2$\,{\scriptsize$[-0.1,\,+4.6]$} & $0.0$\,{\scriptsize$[-1.6,\,+1.7]$} & 0.80\\
OpenCLIP B/16 & $\mathbf{+3.8}$\,{\scriptsize$[+0.1,\,+8.3]$} & $-3.1$\,{\scriptsize$[-7.2,\,+1.5]$} & $\mathbf{-4.3}$\,{\scriptsize$[-7.5,\,-1.4]$} & $-0.5$\,{\scriptsize$[-2.5,\,+1.7]$} & $-1.1$\,{\scriptsize$[-2.6,\,+0.4]$} & 0.73\\
SigLIP B/16 & $+1.2$\,{\scriptsize$[-2.2,\,+5.2]$} & $\mathbf{-5.4}$\,{\scriptsize$[-8.8,\,-2.5]$} & $-0.5$\,{\scriptsize$[-3.1,\,+2.3]$} & $+1.0$\,{\scriptsize$[-0.7,\,+2.8]$} & $-0.3$\,{\scriptsize$[-1.5,\,+0.9]$} & 0.77\\
EVA02-CLIP B/16 & $-3.3$\,{\scriptsize$[-8.1,\,+0.5]$} & $-0.6$\,{\scriptsize$[-4.3,\,+3.2]$} & $+0.7$\,{\scriptsize$[-1.6,\,+3.1]$} & $\mathbf{+3.0}$\,{\scriptsize$[+0.8,\,+5.5]$} & $+0.9$\,{\scriptsize$[-0.5,\,+2.4]$} & 0.58\\
CLIP L/14 & $-1.5$\,{\scriptsize$[-7.3,\,+4.5]$} & $-2.1$\,{\scriptsize$[-5.9,\,+1.4]$} & $-3.0$\,{\scriptsize$[-7.6,\,+1.7]$} & $+1.7$\,{\scriptsize$[-0.4,\,+3.9]$} & $+0.1$\,{\scriptsize$[-1.6,\,+1.7]$} & 0.45\\
\bottomrule
\end{tabular}\end{adjustbox}
\caption{\textbf{Extensions across backbones} (top-1 points, seed 0; bold: interval excludes zero; $I^2$: heterogeneity across the four sets). Fused hard linking is below CLIP-memory hard linking in every pair. The fused gated variant is mostly below Ours and beats it significantly only once (CLIP L/14, Tiny-IN). Transduction alone helps in 19 of 20 pairs, while transductive link re-estimation on top of it changes sign across sets and is never significant after pooling.}
\label{tab:bbext}
\end{table*}

\begin{table*}[h!]
\centering\small
\setlength{\tabcolsep}{4pt}
\begin{adjustbox}{max width=\textwidth}\begin{tabular}{@{}llcccccccc@{}}
\toprule
Test set & Seed & $n$ & Gap & Ours $-$ ens. & Gate effect & Oracle $-$ raw & Fused hard $-$ hard & Transd.\ on ens. & Link EM $-$ transd.\ ens.\\
\midrule
\multirow{2}{*}{Tiny-IN} & 0 & 65 & 14.8 & $+0.6$\,{\scriptsize$[-1.3,\,+2.8]$} & $\mathbf{+2.8}$\,{\scriptsize$[+0.2,\,+5.6]$} & $\mathbf{+12.7}$\,{\scriptsize$[+6.1,\,+20.4]$} & $\mathbf{-8.1}$\,{\scriptsize$[-14.7,\,-1.2]$} & $\mathbf{+2.8}$\,{\scriptsize$[+1.5,\,+4.1]$} & $+3.3$\,{\scriptsize$[-0.3,\,+7.7]$}\\
 & 1 & 67 & 10.6 & $+0.4$\,{\scriptsize$[-1.5,\,+2.1]$} & $\mathbf{+2.5}$\,{\scriptsize$[+1.1,\,+4.0]$} & $\mathbf{+10.2}$\,{\scriptsize$[+3.8,\,+17.1]$} & $\mathbf{-7.7}$\,{\scriptsize$[-13.8,\,-2.0]$} & $\mathbf{+3.1}$\,{\scriptsize$[+1.3,\,+5.2]$} & $-1.5$\,{\scriptsize$[-6.1,\,+3.6]$}\\
\midrule
\multirow{2}{*}{IN-R} & 0 & 68 & 16.5 & $+1.9$\,{\scriptsize$[-0.1,\,+4.0]$} & $\mathbf{+2.4}$\,{\scriptsize$[+0.8,\,+4.3]$} & $\mathbf{+13.1}$\,{\scriptsize$[+7.5,\,+19.1]$} & $\mathbf{-10.2}$\,{\scriptsize$[-16.7,\,-4.5]$} & $\mathbf{+1.8}$\,{\scriptsize$[+0.1,\,+3.5]$} & $\mathbf{-4.2}$\,{\scriptsize$[-8.4,\,-0.9]$}\\
 & 1 & 65 & 11.9 & $+0.3$\,{\scriptsize$[-1.2,\,+1.8]$} & $\mathbf{+2.4}$\,{\scriptsize$[+0.7,\,+4.6]$} & $\mathbf{+7.6}$\,{\scriptsize$[+3.3,\,+12.6]$} & $\mathbf{-8.6}$\,{\scriptsize$[-13.8,\,-4.0]$} & $\mathbf{+2.9}$\,{\scriptsize$[+1.4,\,+4.4]$} & $\mathbf{-5.7}$\,{\scriptsize$[-9.4,\,-2.4]$}\\
\midrule
\multirow{2}{*}{IN-A} & 0 & 66 & 17.6 & $\mathbf{+2.3}$\,{\scriptsize$[+0.2,\,+4.6]$} & $+1.8$\,{\scriptsize$[-1.1,\,+4.7]$} & $\mathbf{+17.1}$\,{\scriptsize$[+11.6,\,+22.8]$} & $-4.5$\,{\scriptsize$[-9.7,\,+0.2]$} & $\mathbf{+4.7}$\,{\scriptsize$[+2.3,\,+7.1]$} & $-3.7$\,{\scriptsize$[-7.2,\,0.0]$}\\
 & 1 & 64 & 15.1 & $+0.1$\,{\scriptsize$[-2.6,\,+3.2]$} & $-0.5$\,{\scriptsize$[-2.7,\,+1.6]$} & $\mathbf{+12.4}$\,{\scriptsize$[+6.9,\,+18.2]$} & $\mathbf{-6.4}$\,{\scriptsize$[-12.1,\,-0.8]$} & $+1.5$\,{\scriptsize$[-1.5,\,+4.7]$} & $+1.2$\,{\scriptsize$[-2.8,\,+5.5]$}\\
\midrule
\multirow{2}{*}{IN-V2} & 0 & 294 & 14.9 & $\mathbf{+2.2}$\,{\scriptsize$[+0.7,\,+3.6]$} & $\mathbf{+5.2}$\,{\scriptsize$[+3.9,\,+6.6]$} & $\mathbf{+14.9}$\,{\scriptsize$[+10.9,\,+19.0]$} & $\mathbf{-6.0}$\,{\scriptsize$[-9.2,\,-3.1]$} & $\mathbf{+1.9}$\,{\scriptsize$[+0.5,\,+3.4]$} & $+2.2$\,{\scriptsize$[-0.1,\,+4.6]$}\\
 & 1 & 292 & 16.7 & $\mathbf{+2.1}$\,{\scriptsize$[+1.1,\,+3.3]$} & $\mathbf{+3.7}$\,{\scriptsize$[+2.0,\,+5.4]$} & $\mathbf{+15.6}$\,{\scriptsize$[+11.6,\,+19.7]$} & $\mathbf{-5.5}$\,{\scriptsize$[-8.3,\,-2.7]$} & $\mathbf{+1.9}$\,{\scriptsize$[+0.9,\,+2.9]$} & $\mathbf{+2.9}$\,{\scriptsize$[+0.6,\,+5.2]$}\\
\bottomrule
\end{tabular}\end{adjustbox}
\caption{\textbf{Second seed} (CLIP B/16). The seed changes the class split, the held-out names and the vocabulary draws, so the two rows of a set are different test problems, not repeated measurements of one. $n$: test classes; Gap: renaming cost (points). Paired top-1 differences with 95\% class-level intervals; bold excludes zero.}
\label{tab:seed}
\end{table*}

\paragraph{Reproducibility.} The main results come from notebook v7 (seed 0) with OpenAI CLIP ViT-B/16 (QuickGELU) and sentence-transformers all-mpnet-base-v2. The backbone study comes from notebook v9, which runs the same cells for each backbone and test set. Its CLIP B/16 seed-0 runs reproduce every v7 training-free number we compared to four decimals: 1312 values, covering the 15 arms' metrics, the paired intervals, and the extra diagnostics on all four sets. The link calibration of Sec.~\ref{sec:ext} is re-selected per backbone and set; it chose $\beta\in\{0.25,0.5,0.75\}$ and $\delta\in\{0.5,1.0\}$. The calibration strings are 3000 held-out views of non-class physical-entity concepts, drawn from the 11{,}329--11{,}662 available per set. Gloss scrubbing removed 50{,}348--54{,}943 n-grams per set.

\end{document}